%% file: main.tex
\documentclass[onecolumn,draftclsnofoot,letterpaper]{IEEEtran}
\IEEEoverridecommandlockouts

\usepackage[utf8]{inputenc} 
\usepackage[T1]{fontenc}
\usepackage{url}
\usepackage{tikz}
\tikzstyle{arrow} = [thick,->,>=stealth]
\usetikzlibrary{decorations.pathreplacing}
\usepackage[cmex10]{amsmath}
\usepackage{amsthm}
\usepackage{amssymb,amsfonts,dsfont,bm,mathrsfs}
\usepackage{siunitx}
\usepackage[bookmarksopen=true,pagebackref=true]{hyperref}
\usepackage{cleveref}
\usepackage[noadjust]{cite}
\usepackage{graphicx}
\allowdisplaybreaks[2]
\usepackage{stfloats}
\usepackage{balance}
\usepackage{enumerate}
\usepackage{makecell}
\usepackage{multirow}
\usepackage{empheq}

{
\theoremstyle{plain}
\newtheorem{theorem}{Theorem}
\newtheorem{proposition}{Proposition}
\newtheorem{lemma}{Lemma}

\theoremstyle{definition}

\theoremstyle{remark}
\newtheorem{remark}{Remark}
}

\begin{document}

\title{Adaptive Refinement Protocols for Distributed Distribution Estimation under $\ell^p$-Losses}

%

\author{Deheng Yuan, Tao Guo and  Zhongyi Huang
\thanks{Deheng Yuan and Zhongyi Huang are with the Department of Mathematical Sciences, Tsinghua University, Beijing 100084, China (emails: ydh22@mails.tsinghua.edu.cn, zhongyih@tsinghua.edu.cn).

Tao Guo is with the School of Cyber Science and Engineering, Southeast University, Nanjing 210096, China (email: taoguo@seu.edu.cn). 
}
}
\maketitle

\begin{abstract}
Consider the communication-constrained estimation of discrete distributions under $\ell^p$ losses, where each distributed terminal holds multiple independent samples and uses limited number of bits to describe the samples. 
We obtain the minimax optimal rates of the problem in most parameter regimes. 
An elbow effect of the optimal rates at $p=2$ is clearly identified. 
To show the optimal rates, we first design estimation protocols to achieve them.
The key ingredient of these protocols is to introduce adaptive refinement mechanisms, which first generate rough estimate by partial information and then establish  refined estimate in subsequent steps guided by the rough estimate.
The protocols leverage successive refinement, sample compression, thresholding and random hashing methods to achieve the optimal rates in different parameter regimes.
The optimality of the protocols is shown by deriving compatible minimax lower bounds.
\end{abstract}


\input{1}

\input{2}

\input{3}

\input{4}

\input{5}
\input{6}

\input{7}

\input{8}
\input{9}

\input{10}

\input{A1}

\input{A2}

\input{A3}

\input{A4}
\input{A5}
\input{A6}
\input{A7}





\bibliographystyle{bibliography/IEEEtran}
\bibliography{bibliography/computing}

\end{document}

%% file: 1.tex
\section{Introduction}
\label{sec:introduction}

Motivated by applications in areas such as federated learning~\cite{McMahan2017,Li2020,Kairouz2021},
distributed statistical estimation problems have recently received wide attention.
In this setting, multiple distributed agents cooperate to train a model, while each of them can only access to a subset of training data.
These agents can exchange messages but their communication budgets are constrained.
The performance of the system is often limited by the communication constraints.

One fundamental learning task is to estimate the underlying discrete distribution of the data. 
Under communication constraints, the minimax optimal rates for the estimation error were studied in~\cite{Diakonikolas2017,Acharya2019,Chen2020,Barnes2020,Han2021,Chen2021,Chen2021point}. 
Another important constraint is the differential privacy, and the corresponding problem was similarly considered in ~\cite{Kairouz2016,Ye2018,Acharya2019,Chen2020}.
In these works, $n = 1$ sample was accessed by each distributed terminal and the most common $\ell^1$ and $\ell^2$ losses were used to measure the estimation error.
However, this is an oversimplification of the practical case, where
general $\ell^p$ losses may be necessary and each terminal can access to  $n>1$ samples. 

On the one hand, \cite{Acharya2021,Acharya2023b} further explored  the distribution estimation problem 
 with $n>1$ samples at each terminal, under the $\ell^1$ loss.
On the other hand, \cite{Acharya2023a,Chen2024} considered the problem under general $\ell^p$ losses, with a  limited scope to $n = 1$. 
In the more practical case where each terminal can obtain $n>1$ samples, the optimal rates under $\ell^p$ losses are still unclear.
The problem with $n>1$ samples is much more difficult than that for $n = 1$, since its inherent structure is not revealed in the $n = 1$ case.
Even though \cite{Acharya2021} presented an optimal protocol for $n>1$ and the $\ell^1$ loss, it still does not directly apply to $\ell^p$ losses since its optimality depends heavily on several special properties of the $\ell^1$ loss.

In this work, we consider the distributed estimation of discrete distributions under communication constraints. The range of the problem is expanded in two directions,  
letting each terminal hold~$n>1$ samples and imposing general $\ell^p$ losses simultaneously.
We design interactive protocols to achieve optimal rates in this technically more challenging setting.
The  difficulty lies in resource allocation, that is allocating multiple terminals and their communication budgets to the estimation tasks of different distribution entries.
The convergence rate under the $\ell^p$ loss is not optimal for uniform allocation, 
hence resources (i.e. the terminals and their communication budgets) should be invested based on the distribution.
As a result, existing protocols fail to handle the general $\ell^p$ loss with the $n$ samples. 
Instead, we design adaptive refinement mechanisms in the protocol, which obtains rough estimate based on the partial information transmitted by a portion of resources, and uses it to allocate the remaining resources for refining the estimate.

Based on the adaptive refinement mechanisms, we design protocols for different parameter regimes by introducing additional  auxiliary estimation methods, from which upper bounds for the optimal rates are induced. 
We also derive compatible lower bounds for most parameter regimes. Hence the optimality of the protocols is shown and the optimal rates are obtained in these regimes.

\begin{itemize}
\item Motivated by the protocol in~\cite{Acharya2021} for the $\ell^1$ loss, we exploit the classic divide-and-conquer strategy and design a successive refinement estimation protocol equipped with an adaptive resource allocation mechanism. 
The distribution is divided into blocks.
The estimation task is achieved by first estimating the block distribution and then conditional distribution over each block.
In the latter phase, terminals are allocated to estimating the conditional distribution based on the block distribution estimated by the former phase.
The block distribution has a lower dimension, and the divide-and-conquer procedure is not stopped until it is more efficient to estimate each entry directly.
The resulting successive refinement protocol achieves the optimal rates up to logarithmic factors for most parameter regimes with $1 \leq p \leq 2$.
As a by-product, our protocol for $p = 1$ achieves the optimal rates for a larger range of regimes than that in~\cite{Acharya2021}.

\item For $p>2$, we introduce additional sample compression methods to aid the adaptive refinement procedure. The methods compress the description for samples and reduce the communication budget, allowing more samples to be transmitted within  limited budget.
The resulting protocols can achieve the optimal rates for relatively large~$n$.
For $n = 1$, an optimal non-interactive protocol can be designed, by exploiting random hash functions in the protocol. 
To show the optimality, we further establish a compatible lower bound that is strictly better than that in~\cite{Acharya2023a,Chen2024},

\item
The above protocols are not optimal in the regime where the total communication budget is extremely tight.
To the best of our knowledge, the regime has not been discussed in any previous work. 
We resolve it by incorporating a thresholding method into the adaptive refinement procedure. 
\end{itemize}

The expression of the optimal rates under $\ell^p$ losses 
reveals an elbow effect at $p = 2$, providing more insights into the distributed estimation problem. 
It is interesting to compare our results with the elbow effect discovered in the nonparamentric density estimation problem~\cite{Butucea2020,Acharya2024}.
The similarity shows how the optimal rates are affected by the relation between the imposed loss function and the constraints on the estimated object. 

The remaining part of this work is organized as follows. First, the problem is formulated in~\Cref{sec:formulation}.
Then we present the main results in~\Cref{sec:main}. 
We design estimation protocols and prove the upper bound for different parameter regimes in~\Cref{sec:base,sec:successive,sec:multiplesample,sec:tightbudget,sec:onesample}.
Next the lower bound is derived in~\Cref{sec:lowerbound}.
Finally, a few remarks are given in~\Cref{sec:discussion}.
See~\Cref{subsec:organization} for detailed organization of the technical parts~\Cref{sec:base,sec:successive,sec:multiplesample,sec:tightbudget,sec:onesample,sec:lowerbound}.

%% file: 2.tex
\section{Problem Formulation}
\label{sec:formulation}
Denote a discrete random variable by a capital letter and its finite alphabet by the corresponding calligraphic letter, e.g., $W\in\mathcal{W}$.
We use the superscript $n$ to denote an $n$-sequence, e.g., $W^n=(W_{i})_{i = 1}^n$.
For a finite set $\mathcal{W}$ of size $k = |\mathcal{W}|$, let $\Delta_{\mathcal{W}}$ be the set all the probability measures over $\mathcal{W}$, i.e. $\Delta_{\mathcal{W}} \triangleq \{p(\cdot): p(w) \in [0,1], \forall w \in \mathcal{W}, \sum_{w} p(w) = 1\}$.
Let $\Delta'_{\mathcal{W}}$ be the set of subprobability measures, i.e. $\Delta_{\mathcal{W}}' \triangleq \{p(\cdot): p(w) \in [0,1], \forall w \in \mathcal{W}, \sum_{w} p(w) \leq 1\}$.

Suppose that we want to estimate the finite-dimensional distribution $p_W \in \Delta_{\mathcal{W}}$ with dimension $k$, and the samples are generated at random. 
To be precise, let $W_{ij} \sim p_{W}(w), i=1,2,\cdots,m$, $j=1,2,\cdots,n$ be i.i.d. random variables distributed over $\mathcal{W}$.
The total sample size is $mn$.
 
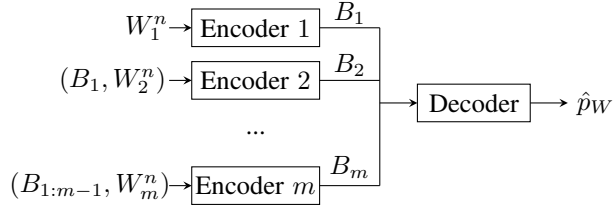
\begin{figure}[!t]
\centering
\begin{tikzpicture}
	\node at (0.9,1) {$W_1^n$};
	\draw[->,>=stealth] (1.2,1)--(1.5,1);
	\node at (2.35,1.0) {Encoder $1$};
	\draw (1.5,0.75) rectangle (3.2,1.25);
	\draw (3.2,1)--(4.0,1);
	\node at (3.6,1.2) {$B_1$};

    \node at (0.45,0.3) {$(B_1,W_2^n)$};
	\draw[->,>=stealth] (1.2,0.3)--(1.5,0.3);
	\node at (2.35,0.3) {Encoder $2$};
	\draw (1.5,0.05) rectangle (3.2,0.55);
	\draw (3.2,0.3)--(4.0,0.3);
	\node at (3.6,0.5) {$B_2$};

        \node at (2.35,-0.4) {...};

    \node at (0.15,-1.1) {$(B_{1:m-1},W_m^n)$};
	\draw[->,>=stealth] (1.2,-1.1)--(1.5,-1.1);
	\node at (2.35,-1.1) {Encoder $m$};
	\draw (1.5,-1.35) rectangle (3.2,-0.85);
	\draw (3.2,-1.1)--(4.0,-1.1);
	\node at (3.6,-0.85) {$B_m$};

    \draw (4.0,1)--(4.0,-1.1);
    \draw[->,>=stealth] (4.0,0)--(4.5,0);
 
	\node at (5.25,0) {Decoder};
	\draw (4.5,-0.25) rectangle (6,0.25);
	
	\draw[->,>=stealth] (6,0)--(6.5,0);
	\node [right] at (6.5,0) {$\hat{p}_W$};
\end{tikzpicture}
\caption{Distributed (sequentially) interactive distribution estimation}
\label{fig:system_model1}
\vspace{-0.2cm}
\end{figure}

Consider the distributed minimax parametric distribution estimation problem with communication constraints depicted in Fig.~\ref{fig:system_model1}.
There are $m$ encoders and one decoder, and common randomness is shared among them. The $i$-th encoder observes the samples $W_i^n = (W_{ij})_{j = 1}^n$ and transmits an encoded message $B_i$ of length $l$ to the decoder, $i =1,...,m$.
Upon receiving messages $B^m = (B_i)_{i = 1}^m$, the decoder needs to establish a reconstruction $\hat{p}_W \in \Delta_{\mathcal{W}}'$ of $p_W$.

An $(m,n,k,l)$-protocol $\mathcal{P}$ is defined by a series of random encoding functions 
\begin{align*}
\mathrm{Enc}_i: \mathcal{W}^n  \times \{0,1\}^{(i-1)l} \to \{0,1\}^{l}, \forall i =1,...,m,
\end{align*}
and a random decoding function 
\[
\mathrm{Dec}:\{0,1\}^{ml} \to \Delta_{\mathcal{W}}'.
\]
The $i$-th encoder is aware of the messages sent by the previous $i-1$ encoders (which can be achieved by interacting with other encoders and/or the decoder), and it generates a binary sequence $B_i = \mathrm{Enc}_i(X^n,B_{1:i-1})$.
The reconstruction of the distribution is $\hat{\bm{p}}_W^{\mathcal{P}} = \mathrm{Dec}(B_{1},B_{2},...,B_{m})$.

For $p\geq 1$, we use the $\ell^p$ loss to measure the estimation error. 
We are interested in the minimal error of all the estimation protocols in the worst case, as the true distribution $p_W$ varies in the probability simplex $\Delta_\mathcal{W}$. 
To be specific, our goal is to characterize the order of the the following minimax convergence rate
\begin{equation*}
    R(m,n,k,l,p) = \inf_{\text{$(m,n,k,l)$-protocol $\mathcal{P}$}} \sup_{\bm{p}_W \in \Delta_{\mathcal{W}}} \mathbb{E}[\Vert \hat{\bm{p}}_W^{\mathcal{P}}-\bm{p}_W \Vert_p^p].
\end{equation*}

\begin{remark}
    The $(m,n,k,l)$-protocol $\mathcal{P}$ defined in this work is usually called the (sequentially) interactive protocol in the literature.
    The protocol is called 
    non-interactive, if for each  $i=1,...,m$, the $i$-th encoder is ignorant of all the messages~$B_{1:i-1}$ sent by previous encoders and the encoding function $\mathrm{Enc}_i(W^n)$ is a function of the samples only. In most cases we design interactive protocols since it is too hard to construct a non-interactive protocol. For some simple special cases, non-interactive protocol achieving the optimal rates can be constructed, which  will be indicated.
\end{remark}

We further define some necessary notations. For any positive $a$ and $b$, we say $a\preceq b$ if $a \leq c \cdot b$ for some positive constant $c>0$ independent of parameters we are concerned, which should be clear in the context.
The notation $\succeq$ is defined similarly. Then we denote by $a \asymp b$ if both $a \preceq b$ and $a \succeq b$ hold.
Denote by $a \wedge b$ the minimum of two real numbers $a$ and $b$, and $a \vee b$ the maximum.

%% file: 3.tex
\section{Main Results and Our Methods}
\label{sec:main}

\subsection{Optimal Rates for $1 \leq p \leq 2$}
\label{subsec:lessthan2}
First assume that $1 \leq p \leq 2$.
We present the upper bound in the following theorem.

\begin{theorem}
\label{thm:upperboundless}
Let $1 \leq p \leq 2$, then we have
\begin{subequations}
\label{eq:upperboundless}
\begin{empheq}[left ={R(m,n,k,l,p)\preceq}  \empheqlbrace]{align}
        &\frac{k}{(mnl)^{\frac{p}{2}}} \vee \frac{k^{1-\frac{p}{2}}}{(mn)^{\frac{p}{2}}}, &n \geq k ,\  &m(l \wedge k) > 1000 k \log (mn) \log n, \label{eq:upperboundless1}
        \\
        &\frac{k^{1-\frac{p}{2}} \log^{\frac{p}{2}}(\frac{k}{n}+1)}{(ml)^{\frac{p}{2}}} \vee \frac{k^{1-\frac{p}{2}}}{(mn)^{\frac{p}{2}}}, &\frac{k}{2^l} \leq n < k,\  &m(l \wedge n) > 2000 n \log (mn) \log n, \label{eq:upperboundless2}
        \\
        &\frac{k}{(mn2^l)^{\frac{p}{2}}},  &n < \frac{k}{2^l},\  &m(l \wedge n)  > 4000 n \log (mn) \log n,\label{eq:upperboundless3} 
        \\
        &\frac{1}{(ml)^{p-1}},  &\log k <l < n,\  & ml < k. \label{eq:upperboundless4}
\end{empheq}
\end{subequations}
\end{theorem}

\begin{IEEEproof}
The case~\eqref{eq:upperboundless1} is by~\Cref{lem:secbasemain} in~\Cref{sec:base}, cases~\eqref{eq:upperboundless2} and~\eqref{eq:upperboundless3} are by~\Cref{lem:secsuccessivemain} in~\Cref{sec:successive},  and the case~\eqref{eq:upperboundless4} is by~\Cref{lem:sectightbudgetmain} in~\Cref{sec:tightbudget}. We sketch the proof here and details can be found in latter sections.

The upper bound for the first three cases~\eqref{eq:upperboundless1}, \eqref{eq:upperboundless2} and~\eqref{eq:upperboundless3} are by the successive refinement protocol with an adaptive resource allocation mechanism detailed in~\Cref{sec:base,sec:successive}.
The idea can be summarized as the following inductive procedure to estimate the distribution. Assume that $\mathcal{W}$ is divided into blocks, and each block is of size at most~$2^l-1$. First suppose that the distribution $p_B$ of blocks has been estimated to some accuracy. Then each encoder can use its $l$-bit message to describe its samples on a predetermined block. Based on these messages, the decoder then estimates the conditional distribution $p_s$ on the $s$-th block. Combining $p_B$ and $p_s$ for each block~$s$, an estimate of $p_W$ can be immediately obtained. Note that the estimation of $p_W$ relies on the estimation of a distribution $p_B$ with a smaller support. Fewer encoders are needed to for the smaller problem. Once the base case of $k<n$ is estimated, $p_W$ can be refined from these layered block distributions successively.

The final case~\eqref{eq:upperboundless4} is proved with the help of a thresholding method. The idea is that under the extremely tight communication budget, approximating those $p_W(w) \preceq \frac{1}{ml}$ simply by $0$ is better than estimating them. Detailed analysis can be found in~\Cref{subsec:thresholding}.
\end{IEEEproof}

The lower bound in the following lemma under the $\ell^p$ loss can be derived from existing results in~\cite{Acharya2021} under the $\ell^1$ loss, which provides a baseline. The proof can be found in~\Cref{sec:lowerbound}.

\begin{lemma}
\label{lem:lowerboundless}
    For $1 \leq p \leq 2$, we have
\begin{equation*}
    R(m,n,k,l,p) \succeq \left\{
    \begin{aligned}
        &\frac{k}{(mnl)^{\frac{p}{2}}} \vee \frac{k^{1-\frac{p}{2}}}{(mn)^{\frac{p}{2}}}, &n \geq k \log k,\  &m > \left(\frac{k}{l}\right)^2\\
        &\frac{k^{1-\frac{p}{2}}}{(ml\log k)^{\frac{p}{2}}} \vee \frac{k^{1-\frac{p}{2}}}{(mn)^{\frac{p}{2}}}, &\frac{k}{2^l} \leq n < k \log k,\  &m > \left(\frac{k}{l}\right)^2,  \\
        &\frac{k}{(mn2^l)^{\frac{p}{2}}},  &n < \frac{k}{2^l}, \  &mn2^l > k^2, \\
        &\frac{1}{(ml)^{p-1}} \vee \frac{k^{1-\frac{p}{2}}}{(mn)^{\frac{p}{2}}},  & ml < \frac{k}{2}. 
    \end{aligned}\right.
\end{equation*}
\end{lemma}

Combining~\Cref{thm:upperboundless,lem:lowerboundless}, the optimal rates for the following cases can be roughly characterized by
\begin{equation}
\label{eq:optimalrateless2}
    R(m,n,k,l,p) \asymp \left\{
    \begin{aligned}
        &\frac{k}{(mnl)^{\frac{p}{2}}} \vee \frac{k^{1-\frac{p}{2}}}{(mn)^{\frac{p}{2}}}, &n \geq k ,\  &ml \succeq k\\
        &\frac{k^{1-\frac{p}{2}}}{(ml)^{\frac{p}{2}}} \vee \frac{k^{1-\frac{p}{2}}}{(mn)^{\frac{p}{2}}}, &\frac{k}{2^l} \leq n < k,\  &ml \succeq k,  \\
        &\frac{k}{(mn2^l)^{\frac{p}{2}}},  &n < \frac{k}{2^l}, \  &mn2^l \succeq k^2, \\
        &\frac{1}{(ml)^{p-1}} \vee \frac{k^{1-\frac{p}{2}}}{(mn)^{\frac{p}{2}}},  & ml \preceq k. 
    \end{aligned}\right.
\end{equation}

\begin{remark}[About the boundaries in~\eqref{eq:optimalrateless2}]
    We believe that the regularity condition $m>(\frac{k}{l})^2$ in the lower bound is induced mainly by technical reasons and the boundary $ml>k$ is more essential. Similarly, the conditions $m(l \wedge k)>1000k \log (mn) \log n$ and $m(l \wedge n) > 2000n \log (mn) \log n$ in the upper bound can be relaxed by finer analysis and the true boundaries seem to be around $ml>k$ and $ml>n$. Under these observations, in the third case the conditions $mn2^l \geq k^2$ and $n < \frac{k}{2^l}$ imply that $m > k > n$ and hence $ml > n$ is fullfilled.
\end{remark}

\subsection{Optimal Rates for $p = 1$ and $p = 2$}

In this subsection, we specialize our results and characterize the optimal rates under the most commonly used total variation (TV) and squared losses, i.e. $\ell^1$ and $\ell^2$ losses. For the TV loss, the successive refinement protocol can be made non-interactive.
See Appendix~\ref{sec:tvprotocol} for details.

\begin{theorem}
\label{thm:upperboundTV}
The following upper bound can be achieved by a non-interactive protocol.
\begin{equation*}
    R(m,n,k,l,1) \preceq \left\{
    \begin{aligned}
        &\sqrt{\frac{k^2}{mnl}} \vee \sqrt{\frac{k}{mn}}, &n \geq k ,\  &m(l \wedge k) > 1000 k \log m \log n,\\
        &\sqrt{\frac{k\log(\frac{k}{n}+1)}{ml}} \vee \sqrt{\frac{k}{mn}}, &\frac{k}{2^l} \leq n < k,\  &m(l \wedge n) > 2000 n \log m \log n,  \\
        &\sqrt{\frac{k^2}{mn2^l}},  &n < \frac{k}{2^l},\  &m(l \wedge n) > 4000 n \log m \log n. \\ 
    \end{aligned}\right.
\end{equation*}
\end{theorem}

For the TV loss, we have the following characterization of the optimal rates.
    
\begin{equation}
\label{eq:optimalrate1}
R(m,n,k,l,p = 1) \asymp \left\{
\begin{aligned}
    & \sqrt{\frac{k^2}{mnl}} \vee \sqrt{\frac{k}{mn}}, &n \geq k, ml \succeq k, 
    \\
    & \sqrt{\frac{k}{ml}} \vee \sqrt{\frac{k}{mn}}, & \frac{k}{2^l} \leq n < k, \ ml \succeq k, 
    \\
    &\sqrt{\frac{k^2}{mn2^l}} \wedge 1, & n < \frac{k}{2^l},
    \\
    & 1, & ml \preceq k.
\end{aligned}
\right.
\end{equation}

\begin{remark}
The same as~\Cref{thm:upperboundTV}, the non-iteractive protocol in~\cite{Acharya2021} is constructed for the estimation problem under the TV loss.
However, corresponding to the third case in~\Cref{thm:upperboundTV}, 
in~\cite{Acharya2021} a stronger restriction $m> 100\frac{k}{2^l}  \log m \log n$ is imposed (cf. Theorem 1.1 in~\cite{Acharya2021} and note that the notations $m$ and $n$ are interchanged therein).
The restriction is induced by using the first  bit of each encoder to estimate the block probability $p_B$ with the protocol for the first case. The conditional probability in each block $B$ is then estimated. Combining it with the estimate for $p_B$, an estimate for $p_W$ is obtained.
In fact, it is a one-step reduction.
We note that the step that estimates the conditional probability can be abstracted and summarized as a separate protocol, and it has an inductive nature.
Instead of using it only once, we iteratively use the protocol, which is inspired by the classic divide-and-conquer strategy.
Thus our successive refinement protocol relaxes the restriction in~\cite{Acharya2021} and achieve an upper bound for a wider parametric range. 
\end{remark}

The squared loss is the most widely used loss, in both theoretical analysis and algorithm research.
By specializing~\eqref{eq:optimalrateless2}, we have a more complete characterization of the order of $R(m,n,k,l,p = 2)$.  

\begin{equation}
\label{eq:optimalrate2}
R(m,n,k,l,p = 2) \asymp \left\{
\begin{aligned}
    & \frac{k}{mnl} \vee \frac{1}{mn}, & n \geq k, ml \succeq k,
    \\
    & \frac{1}{ml} \vee \frac{1}{mn}, & \frac{k}{2^l} \leq n < k \text{ or } n \geq k, ml \preceq k.
    \\
    &\frac{k}{mn2^l}, & n < \frac{k}{2^l}, mn2^l \succeq k^2,
\end{aligned}
\right.
\end{equation}

\subsection{Optimal Rates for $p>2$}
\label{subsec:greaterthan2}

For $p > 2$, we first present the upper bound in the following.

\begin{theorem}
\label{thm:upperboundgreater}
Let $p > 2$, then we have
\begin{subequations}
\label{eq:upperboundgreater}
\begin{empheq}[left = {R(m,n,k,l,p) \preceq}\empheqlbrace]{align}
        &\frac{k}{(mnl)^{\frac{p}{2}}} \vee \frac{1}{(mn)^{\frac{p}{2}}}, &n \geq k ,\  &m(l \wedge k^{\frac{2}{p}}) > 1000 k \log (mn) \log n, \label{eq:upperboundgreater1}
        \\
        &\frac{\log^{\frac{p}{2}}k}{(ml)^{\frac{p}{2}}n^{\frac{p}{2}-1}} \vee \frac{1}{(mn)^{\frac{p}{2}}}, &\frac{k}{(2^l)^{\frac{p}{2}}} \leq n < k,\  &l > \log k,\notag
        \\
        & & &m(l \wedge n^{\frac{2}{p}}) \geq 1000 n \log (mn) \log k, \label{eq:upperboundgreater2}
        \\
        &\left(\frac{k}{mn2^l}\right)^{\frac{p}{2}},  &n < \frac{k}{(2^l)^{\frac{p}{2}}},\  &m(l \wedge n) > 4000 n \log (mn) \log n, \label{eq:upperboundgreater3}
        \\
        &\frac{\log^p k \vee \log^{2p} (mn)}{(ml)^{p-1}} \vee \frac{1}{(mn)^{\frac{p}{2}}},  &\log k < l < n,\  & ml < n. \label{eq:upperboundgreater4}
\end{empheq}
\end{subequations}
\end{theorem}

\begin{IEEEproof}
The case~\eqref{eq:upperboundgreater1} is by~\Cref{lem:secbasemain} in~\Cref{sec:base}, the case~\eqref{eq:upperboundgreater2} is by~\Cref{lem:secmultiplesamplemain} in~\Cref{sec:multiplesample}, the case~\eqref{eq:upperboundgreater3} is by~\Cref{lem:secsuccessivemain} in~\Cref{sec:successive}, and the case~\eqref{eq:upperboundgreater4} is by~\Cref{lem:sectightbudgetmain} in~\Cref{sec:tightbudget}. We sketch the proof here and details can be found in latter sections.

The bounds in~\eqref{eq:upperboundgreater1} and~\eqref{eq:upperboundgreater2}  are achieved by adaptive refinement protocols.
In both cases, a rough estimate is first established, by assigning the first half of all encoders uniformly to estimating each entry $p_W(w)$.
Based on that, the remaining encoders are allocated to refine different entries according to their order. 
For the first bound, a portion of roughly~$p_W(w)$ is allocated to estimate $p_W(w)$.
The spirit of the allocation mechanism is similar to that designed for the pointwise estimation problem~\cite{Chen2021point}.
For the second bound, a sample compression mechanism is used. 
Note that the number of the elements $w$ with $p_W(w) \succeq \frac{1}{n}$ (denote the set containing those elements $w$ by $\mathcal{W}'$) is about $n$. 
Samples are compressed by projecting them to $\mathcal{W}'$, which saves the communication budget.
Hence those $p_W(w) \succeq \frac{1}{n}$ are refined by invoking the protocol for the first case.
See~\Cref{sec:base,sec:multiplesample} for details.

The bound in~\eqref{eq:upperboundgreater3} is a corollary of the successive refinement protocol in~\Cref{sec:successive}.
The bound in~\eqref{eq:upperboundgreater4} is achieved by exploiting both sample compression and thresholding mechanisms, which is proved in~\Cref{subsec:combining}.
\end{IEEEproof}

Similar to~\Cref{subsec:lessthan2}, we present the lower bound as a baseline in the following lemma proved in~\Cref{sec:lowerbound}.

\begin{lemma}
\label{lem:lowerboundgreater}
For $p > 2$, we have
\begin{equation*}
    R(m,n,k,l,p) \succeq \left\{
    \begin{aligned}
        &\frac{k}{(mnl)^{\frac{p}{2}}} \vee \frac{1}{(mn)^{\frac{p}{2}}}, &n \geq k \log k,\  &m > \left(\frac{k}{l}\right)^2\\
        &\frac{1}{(ml)^{\frac{p}{2}}n^{\frac{p}{2}-1}\log n} \vee \frac{1}{(mn)^{\frac{p}{2}}}, &\frac{k}{(2^l)^{\frac{p}{2}}} \leq n < k \log k,\  &m > \left(\frac{n/\log n}{l}\right)^2,  \\
        &\frac{k}{(mn2^l)^{\frac{p}{2}}},  &n< \frac{k}{(2^l)^{\frac{p}{2}}},\ &mn2^l > k^2,\\
        &\frac{1}{(ml)^{p-1}}\vee \frac{1}{(mn)^{\frac{p}{2}}},    & ml < \frac{k}{2}.    
    \end{aligned}\right.
\end{equation*}
\end{lemma}

Combining~\Cref{thm:upperboundgreater,lem:lowerboundgreater}, the optimal rates for the following cases can be roughly characterized.
\begin{equation}
\label{eq:optimalrategreater2}
    R(m,n,k,l,p) \asymp \left\{
    \begin{aligned}
        &\frac{k}{(mnl)^{\frac{p}{2}}} \vee \frac{1}{(mn)^{\frac{p}{2}}}, &n \geq k ,\  ml \succeq n\\
        &\frac{1}{(ml)^{\frac{p}{2}}n^{\frac{p}{2}-1}} \vee \frac{1}{(mn)^{\frac{p}{2}}}, &\frac{k}{(2^l)^{\frac{p}{2}}} \leq n < k,\  ml \succeq n,  \\
        &\frac{k}{(mn2^l)^{\frac{p}{2}}},  &n< \frac{k}{(2^l)^{\frac{p}{2}}},\ mn2^l \succeq k^2,\\
        &\frac{1}{(ml)^{p-1}}\vee \frac{1}{(mn)^{\frac{p}{2}}},    & ml \preceq k, k<n \text{ or } ml \preceq n, k> n.    
    \end{aligned}\right.
\end{equation}

\subsection{Optimal Rates for $p>2$ and $n = 1$}

For $n = 1$ and $p > 2$, the lower bound can be derived by specializing~\Cref{lem:lowerboundgreater}, and the compatible upper bound is achieved by a non-interactive protocol, shown in the following thoerem. 

\begin{theorem}
\label{thm:onesample}
Let $p > 2$ and $n = 1$. 
We can design a non-interactive protocol that achieves the upper bound $R(m,1,k,l,p) \preceq \frac{k}{(m2^l)^{\frac{p}{2}}} \vee \frac{1}{m^{\frac{p}{2}}}$.
If $m(2^l \wedge k^{\frac{2}{p}}) \geq k^2$, then  $R(m,1,k,l,p) \succeq \frac{k}{(m2^l)^{\frac{p}{2}}} \vee \frac{1}{m^{\frac{p}{2}}}$.
\end{theorem}

\begin{IEEEproof}
The upper bound is by~\Cref{lem:seconesamplemain} in~\Cref{sec:onesample}.
We sketch the proof here.
A random hash function is used to compress the sample first, and then the estimate can be directly obtained by constructing and rescaling the histogram.
See~\Cref{sec:onesample} for details of the protocol and its analysis.
\end{IEEEproof}

\begin{remark}
Note that the central bound $\frac{1}{m^{\frac{p}{2}}}$ without the communication constraints is neglected by previous works~\cite{Acharya2023a,Chen2024} (see Theorem~6 in~\cite{Acharya2023a} and Corollary 3.2 in~\cite{Chen2024}). Hence the lower bounds in both works are clearly not tight (for $p>2$). The work~\cite{Chen2024} further claimed that the lower bound $\frac{k}{(m2^l)^{\frac{p}{2}}} \vee \frac{k^{1-\frac{p}{2}}}{m^{\frac{p}{2}}}$ is optimal (see Lemma 3.3 therein), but the sketch given there is too brief and not sufficient to describe a protocol that achieves the bound. In fact, given that the lower bound in~\cite{Chen2024} can be strictly improved, it is impossible to show its optimality. Moreover, constructing the protocol that achieves the optimal rates for $p>2$ is not that straightforward and needs additional ideas.
We use random hashing to resolve the difficulty in this work.

\end{remark}

\subsection{Summary of the Optimal Rates}
\label{subsec:summarymain}
In~\Cref{tab:optimalrate}, we summarize the characterizations of the optimal rate obtained in~\Cref{eq:optimalrate1,eq:optimalrate2,eq:optimalrategreater2,eq:optimalrateless2} and~\Cref{thm:onesample}. 
The essential bounds originally derived in this work are highlighted in red, while those established in previous works~\cite{Acharya2021,Han2021,Barnes2020,Acharya2023a,Chen2024} are shown in blue. 
All the other bounds are corollaries of them.
The optimal rates (up to logarithmic factors) are obtained for most cases, except the case $p > 2$, $n< \frac{k}{(2^l)^{\frac{p}{2}}}$ and $mn2^l \geq k^2$, where our lower and upper bounds do not coincide.
Though a good news is that for its special case $n = 1$, the optimal rates can be obtained. 
We conjecture that the lower bound $\frac{k}{(mn2^l)^{\frac{p}{2}}}$ is tight, which is partially verified in the case $n = 1$. 

We find several interesting phenomena of the optimal rates.
First, note that there is an elbow effect in the parameter~$p$ between the regimes $1 \leq p < 2$ and $p \geq 2$.
The difference is clearly reflected in the central bound without any communication constraints, i.e. $l = \infty$.
The bound is $\frac{k^{1-\frac{p}{2}}}{(mn)^{\frac{p}{2}}}$ for $1 \leq p < 2$, while for $p \geq 2$ it is $\frac{1}{(mn)^{\frac{p}{2}}}$ and independent of the dimension~$k$ of the distribution.
The other sharp difference is that, for a medium $n$, i.e. $\frac{k}{(2^l)^{\frac{p}{2}\vee 1}} \leq n < k$, the optimal rate is independent of $k$ (up to logarithmic factors) for $p \geq 2$, which is not the case for $1 \leq p < 2$.

Second, the minimum transmitted bits required for recovering the same rates in the central case without any communication constraints are interesting for $p>2$. It is roughly $k^{\frac{2}{p}}$ for $k < n$, $ml \geq k$ and $n^{\frac{2}{p}}$ for $k \geq n$, $ml \geq n$, which is out of expectation.
It shows a shrinkage compared to the required number of bits $k$ and $n$ for the case $1 \leq p < 2$.
Similarly, for $n = 1$ and $m2^l \geq k^2$, the required number of bits is roughly $\frac{2}{p} \log k$ instead of $\log k$.

The last observation is that if the total communication budget is extremely tight ($ml \ll k$), then the optimal rates is dependent only on the total budget and independent of the parameters $k$ and $n$. This parameter regime has not been carefully studied in previous work to our best knowledge.

\begin{table}
\caption{Bounds of $R(m,n,k,l,p)$ for Different Cases}
    \centering
    \begin{tabular}{|c|c|c|c|c|}
    \hline
    Parameter Regimes& $p = 1$ & $1 \leq p \leq 2$ & $p = 2$ & $p \geq 2$    
    \\
    \hline
    $l = \infty$ & {\color{blue}$R \asymp \sqrt{\frac{k}{mn}}$} & $R \asymp \frac{k^{1-\frac{p}{2}}}{(mn)^{\frac{p}{2}}}$ & $R \asymp \frac{1}{mn}$ & {\color{red} $R \asymp \frac{1}{(mn)^{\frac{p}{2}}}$ (\Cref{lem:lowerboundcentral})}  
    \\
    \hline
    $n \geq k$, $l^{\frac{p}{2} \vee 1} \leq k$, & \multirow{2}{*}{\color{blue}$R \asymp \frac{k}{\sqrt{mnl}}$} & \multirow{2}{*}{$R \asymp \frac{k}{(mnl)^{\frac{p}{2}}}$} & \multirow{2}{*}{$R \asymp \frac{k}{mnl}$} & \multirow{2}{*}{{\color{red} $R \asymp \frac{k}{(mnl)^{\frac{p}{2}}}$ (\Cref{lem:secbasemain})}}
    \\
    $ml \geq k$& & & &
    \\
    \hline
    $\frac{k}{(2^l)^{\frac{p}{2} \vee 1}} \leq n < k$, $l^{\frac{p}{2} \vee 1} \leq n$, & \multirow{3}{*}{{\color{blue}$R \asymp \sqrt{\frac{k}{ml}}$}} & \multirow{3}{*}{$R \asymp \frac{k^{1-\frac{p}{2}}}{(ml)^{\frac{p}{2}}}$} & \multirow{3}{*}{$R \asymp \frac{1}{ml}$} & \multirow{3}{*}{\color{red} $R \asymp \frac{1}{(ml)^{\frac{p}{2}}n^{\frac{p}{2}-1}}$ }
    \\
    $ml \geq k$ ($p \leq 2$), & & & & 
    \\
    $ml \geq n$ ($p > 2$) & & & &{\color{red}(\Cref{lem:secsuccessivemain,lem:secmultiplesamplemain})}
    \\
    \hline
    $ml < k$ ($p \leq 2$ or $p >2$, $k\leq n$),& \multirow{2}{*}{{\color{blue}$R \asymp 1$}} & \multirow{2}{*}{\color{red} $R \asymp \frac{1}{(ml)^{p-1}}$ (\Cref{lem:sectightbudgetmain})}& \multirow{2}{*}{$R \asymp \frac{1}{ml}$} & \multirow{2}{*}{\color{red}$R \asymp \frac{1}{(ml)^{p-1}}$ (\Cref{lem:sectightbudgetmain})}
    \\
     $ml < n$ ($p > 2$, $k > n$), $l > \log k$& & & &
    \\
    \hline
    $n < \frac{k}{(2^l)^{\frac{p}{2} \vee 1}}$, & \multirow{2}{*}{{\color{blue}$R \asymp \frac{k}{\sqrt{mn2^l}}$ }} & \multirow{2}{*}{$R \asymp \frac{k}{(mn2^l)^{\frac{p}{2}}}$} & \multirow{2}{*}{$R \asymp \frac{k}{mn2^l}$} & {\color{red} $R \preceq \left(\frac{k}{mn2^l}\right)^{\frac{p}{2}}$ (\Cref{lem:secsuccessivemain})}
    \\
    $mn2^l \geq k^2$ & & & &$R \succeq \frac{k}{(mn2^l)^{\frac{p}{2}}}$ 
    \\
    \hline
    $n = 1$, $(2^l)^{\frac{p}{2}\vee 1} < k$, & \multirow{2}{*}{\color{blue}$R \asymp \frac{k}{\sqrt{m2^l}}$} & \multirow{2}{*}{$R \asymp \frac{k}{(m2^l)^{\frac{p}{2}}}$} & \multirow{2}{*}{\color{blue} $R \asymp \frac{k}{m2^l}$} & \multirow{2}{*}{{\color{red} $R \asymp \frac{k}{(m2^l)^{\frac{p}{2}}}$ (\Cref{lem:seconesamplemain})}}
    \\
    $m2^l \geq k^2$ & & & &
    \\
    \hline
    \end{tabular}
    \label{tab:optimalrate}
\end{table}

\subsection{Organization of the Remaining Part of the Work}
\label{subsec:organization}

The remaining part of this work is devoted to presenting the detailed proof of the main results, by designing optimal protocols to achieve the upper bounds for different parameter regimes in~\Cref{sec:base,sec:successive,sec:multiplesample,sec:tightbudget,sec:onesample} and deriving the compatible (up to logarithmic factors) lower bounds in~\Cref{sec:lowerbound}.
These sections are organized as in~\Cref{tab:optimalrate} and follows.
\begin{itemize}
    \item \Cref{sec:base} presents the adaptive refinement protocol for cases~\eqref{eq:upperboundless1} and~\eqref{eq:upperboundgreater1} in~\Cref{thm:upperboundless,thm:upperboundgreater}, summarized in~\Cref{lem:secbasemain}.
    \item \Cref{sec:successive} presents the adaptive successive refinement protocol with resource allocation for cases~\eqref{eq:upperboundless2} and~\eqref{eq:upperboundless3} in~\Cref{thm:upperboundless} and~\eqref{eq:upperboundgreater3} in~\Cref{thm:upperboundgreater}, summarized in~\Cref{lem:secsuccessivemain}.
    \item \Cref{sec:multiplesample} presents the adaptive refinement protocol with sample compression methods for the case~\eqref{eq:upperboundgreater2} in~\Cref{thm:upperboundgreater}, summarized in~\Cref{lem:secmultiplesamplemain}.
    \item \Cref{sec:tightbudget} presents the adaptive refinement protocol with thresholding methods for cases~\eqref{eq:upperboundless4} and~\eqref{eq:upperboundgreater4} in~\Cref{thm:upperboundless,thm:upperboundgreater}, summarized in~\Cref{lem:sectightbudgetmain}.
    \item \Cref{sec:onesample} presents the non-interactive protocol based on random hashing for the $n = 1$ case in~\Cref{thm:onesample}, summarized in~\Cref{lem:seconesamplemain}.
    \item \Cref{sec:lowerbound} shows all the lower bounds in~\Cref{lem:lowerboundless,lem:lowerboundgreater}. 
    \item \Cref{sec:discussion} gives some  further discussions.
\end{itemize}

%% file: 4.tex
\section{The Protocol for Cases~\eqref{eq:upperboundless1} and~\eqref{eq:upperboundgreater1}}
\label{sec:base}

In this section, we design an adaptive refinement protocol that achieves the optimal rates for cases~\eqref{eq:upperboundless1} and~\eqref{eq:upperboundgreater1}, summarized in the following proposition. 

\begin{proposition}
\label{lem:secbasemain}
Let $p \geq 2$, $k \leq n$, $ml > 1000 k \log (mn) \log n$ and $l \leq k^{\frac{2}{p}}$. 
Then for the estimation problem in~\Cref{sec:formulation}, there exists an interactive protocol $\mathrm{AR}(m,n,k,l,p)$ such that for any $\bm{p}_W \in \Delta_{\mathcal{W}}$, the protocol outputs an estimate $\hat{\bm{p}}_W$ satisfying $\mathbb{E}[\Vert \hat{\bm{p}}_W-\bm{p}_W \Vert_p^p] =  O\left(\frac{k}{(mnl)^{\frac{p}{2}}}\right)$.
\end{proposition}

\begin{remark}
\label{rem:Holder}
With the help of~\Cref{lem:secbasemain}, then for $1 \leq p < 2$, let the protocol $\mathrm{AR} (m,n,k,l,p)$ be the same as that for $p = 2$, i.e., $\mathrm{AR} (m,n,k,l,2)$. Then by the H\"{o}lder's inequality, we have
\begin{equation*}
    \mathbb{E}[\Vert\hat{\bm{p}}_W-\bm{p}_W\Vert_p^p] \leq k^{1-\frac{p}{2}}\left(\mathbb{E}[\Vert\hat{\bm{p}}_W-\bm{p}_W\Vert_p^2]\right)^{\frac{p}{2}}.
\end{equation*}
Hence 
\begin{equation*}
    R(m,n,k,l,p) \leq k^{1-\frac{p}{2}} R(m,n,k,l,2)^{\frac{p}{2}},
\end{equation*}
and the minimax upper bound for $1 \leq p < 2$ is easily implied by that for $p = 2$.
\end{remark}

Now return to the proof of~\Cref{lem:secbasemain}.
Each entry of the distribution can be estimated by invoking the one-bit protocol in~\cite{Acharya2021} for the estimation of a binary distribution. We first show the error bound in the following lemma, which can be proved by adapting the proof of Theorem A.2 and A.3 therein. 

\begin{lemma}
\label{lem:binaryestimate}
Suppose that there are $m'$ users and each of them observe an i.i.d. sample from the binary distribution $\mathrm{B}(n,q)$ and $m' > 1000\log n$. Then for $p \geq 2$, there exists a one-bit protocol which outputs an estimate $\hat{q}$ satisfying
\begin{equation}
\mathbb{E}\left[|q-\hat{q}|^p\right] = O\left( \left(\frac{q}{m'n}\right)^{\frac{p}{2}} + \left(\frac{q}{n}\vee \frac{1}{n^2}\right)^{\frac{p}{2}}e^{-\frac{m'}{240 \log n}}\right).
\end{equation}
\end{lemma}

\subsection{The Adaptive Refinement Protocol}

\subsubsection{Rough Estimation}

The first step is to let the first $\frac{m}{2}$ encoders and the decoder jointly generate a rough estimate $\hat{\bm{p}}^1$.
Let  $m' = \lfloor\frac{ml}{2k}\rfloor$. Each encoder can  concurrently run $l$ one-bit protocols in~\Cref{lem:binaryestimate}  using its $l$ bits, where $l \leq k^{\frac{2}{p}} \leq k \leq n$ and the goal of each protocol is to estimate $\bm{p}_W$ for some $w \in \mathcal{W}$. At the same time, a proper allocation plan can ensure that for each $w \in \mathcal{W}$, there are $m'$ encoders running the protocol for estimating $\bm{p}_W$.
The decoder then obtains the rough estimate $\hat{\bm{p}}_W^1$.

\subsubsection{Refinement of the Estimate}

The second step is to let the next $\frac{m}{2}$ encoders and the decoder jointly generate a refined estimate $\hat{\bm{p}}_W^2$.
Let  $m(w) = \lfloor\frac{ml(\hat{p}^1_W(w)+ \frac{1}{k})}{4}\rfloor \wedge \frac{m}{2}$.
Each encoder can  concurrently run $l$ one-bit protocols in~\Cref{lem:binaryestimate}  using its $l$ bits, for estimating some $\bm{p}_W$. At the same time, a proper allocation plan can ensure that for each $w \in \mathcal{W}$, there are $m(w)$ encoders\footnote{One may worry that the estimate $\hat{\bm{p}}_W^1$ may not be normalized. But it does not affect the subsequent steps of using $\hat{\bm{p}}_W^1$ for directing the resource allocation. This can be seen by the following analysis. By the proof of Theorem A.2 in~\cite{Acharya2021} and $n \geq k$, for a constant $C>1$, $\mathbb{P}[\Vert \hat{\bm{p}}_W^1 \Vert_1 \geq C] \leq \sum_{w}\mathbb{P}[ |\hat{p}_W^1(w)-p_W(w)|\geq (C-1)(\frac{1}{n}\vee \sqrt{\frac{p_W(w)}{n}})] \leq k \log n \cdot e^{-\frac{m'}{240 \log n}}$, which is sufficiently small if $ml\gg k \log n \log (mn)$. In the case that $\hat{\bm{p}}_W^1$ is used as a ratio for resource allocation, we can simply divide it by the constant $C$ and then the error analysis is still true. Hence we assume that $\hat{\bm{p}}_W^1$ is normalized and do not point out the difference in similar cases where $\hat{\bm{p}}_W^1$ is generated by the protocol in~\Cref{lem:binaryestimate} for simplicity.} running the protocol for estimating $\bm{p}_W$.
The decoder then constructs the refined estimate $\hat{\bm{p}}_W^2$.

\subsection{Error Analysis}
\label{subsec:baseanalysis}
It is not hard to analyze the error of the rough estimate. 
By~\Cref{lem:binaryestimate} and the assumption $ml >1000 k \log (mn) \log n$, for any $w \in \mathcal{W}$ we have
\begin{align}
\label{eq:crudeerror}
    \mathbb{E}\left[|p_W(w)-\hat{p}^1_W(w)|^p\right] 
    = O\left( \left(\frac{kp_W(w)}{mnl}\right)^{\frac{p}{2}} + \left(\frac{1}{mnl}\right)^{\frac{p}{2}} \right).
\end{align}
However, simply taking the summation can only get the total error bound $O((\frac{k}{mnl})^{\frac{p}{2}})$, which is not tight for $p>2$. 

To obtain the tight bound, our solution is to use the rough estimate $\hat{\bm{p}}_W^1$ for directing the resource allocation in the second step.
Then the refined estimate in the second step can achieve the desired upper bound, i.e. $\mathbb{E}[\Vert\hat{\bm{p}}^2_W-\bm{p}_W\Vert_p^p]  = O\left(\frac{k}{(mnl)^{\frac{p}{2}}}\right)$, which completes the proof of~\Cref{lem:secbasemain}.
See Appendix~\ref{sec:pfbase} for details.

%% file: 5.tex
\section{The Protocol for Cases~\eqref{eq:upperboundless2}, \eqref{eq:upperboundless3} and~\eqref{eq:upperboundgreater3}}
\label{sec:successive}

In this section, we design a successive refinement protocol with adaptive resource allocation that achieves the optimal rates for cases~\eqref{eq:upperboundless2}, \eqref{eq:upperboundless3} and~\eqref{eq:upperboundgreater3}.
Similar to the discussion in~\Cref{rem:Holder}, it suffices to show the following proposition for $p \geq 2$.

\begin{proposition}
\label{lem:secsuccessivemain}
Let $p \geq 2$. Then for the problem in~\Cref{sec:formulation}, there exists an interactive protocol $\mathrm{ASR}(m,n,k,l,p)$ such that for any $\bm{p}_W \in \Delta_{\mathcal{W}}$, the protocol outputs an estimate $\hat{\bm{p}}_W$ satisfying,
\begin{enumerate}[1.]
\item \label{case1} 
if $k \leq n$,  $m(l \wedge k) > 1000 k \log (mn) \log n$, then $\mathbb{E}[\Vert \hat{\bm{p}}_W-\bm{p}_W \Vert_p^p] =  O\left( \left(\frac{k}{mnl}\right)^{\frac{p}{2}} \vee \frac{1}{(mn)^{\frac{p}{2}}}\right)$;

\item \label{case3} if $n < k \leq (2^l-1) \cdot n$, $l \geq 2$ and $m(l \wedge n) >2000  n \log (mn) \log n$, 
then $\mathbb{E}[\Vert \hat{\bm{p}}_W-\bm{p}_W \Vert_p^p] =  O\left( \left(\frac{\log (\frac{k}{n}+1)}{ml} \right)^{\frac{p}{2}}\vee \frac{1}{(mn)^{\frac{p}{2}}} \right)$;

\item \label{case4} if $k >(2^l-1) \cdot n$, $l \geq 4$ and $m(l \wedge n) > 4000 n \log (mn) \log n$, then $\mathbb{E}[\Vert \hat{\bm{p}}_W-\bm{p}_W \Vert_p^p] =  O\left( \left(\frac{k}{2^lmn}\right)^{\frac{p}{2}}\right)$.

\end{enumerate}
\end{proposition}

\begin{remark}
Although the bound in~\Cref{lem:secsuccessivemain} is not always tight for $p>2$, it is indeed tight (up to logarithmic factors) for $p = 2$ and can imply tight bound for $1 \leq p < 2$. 
The advantage of using the successive refinement protocol for $1 \leq p < 2$ is that the protocol can apply for a wider parameter regime. In comparison, the protocol in~\Cref{sec:multiplesample} can be used for $1 \leq p < 2$ and $k>n$ but it requires that $l > \log k$.
Hence it fails to handle the case~\ref{case3} for $\log (\frac{k}{n}+1) < l \leq \log k$ and the case~\ref{case4} in~\Cref{lem:secsuccessivemain}.
\end{remark}

We design the adaptive successve refinement protocol $\mathrm{ASR} (m,n,k,l,p)$ in~\Cref{lem:secsuccessivemain} inductively, which turns out to be a successive refinement procedure.
The protocol for each case in~\Cref{lem:secsuccessivemain} relies on that for preceding cases.
The goal is to estimate a distribution $\bm{p}_W \in \Delta_{\mathcal{W}}$. If the communication budget $l$ for each encoder is too tight, then it is hard to describe all the entries of $\bm{p}_W$. Instead, 
we can perform a a divide-and-conquer strategy.

At each step, choose some $l_0$ and construct a division $\mathcal{W} = \cup_{s = 1}^t \mathcal{W}_s$ with $|\mathcal{W}_s| \leq 2^{l_0}-1$, $l_0 \leq l$ and $t = \lceil \frac{k}{2^{l_0}-1} \rceil$.
Then each encoder is assigned a subset $\mathcal{W}_s$  and ordered to describe the conditional distribution $\bm{p}_s\in \Delta_{\mathcal{W}_s}$, where $p_s(w) \triangleq p(w|\mathcal{W}_s)$. Based on the message, the decoder constructs $\hat{\bm{p}}_s$ as an estimate of $\bm{p}_s$.
Let the block distribution be $\bm{p}_B$, where $p_B(s) = \sum_{w \in \mathcal{W}_s}p(w)$.
If an estimate $\hat{\bm{p}}_B$ of the distribution $\bm{p}_B$ can be obtained, then it is easy to obtain an estimate $p_W(w) = \hat{p}_B(s)\hat{p}_s(w)$ for $w \in \mathcal{W}_s$.

The above procedure can be repeated for the estiamtion of $\bm{p}_B$. 
Note that $\bm{p}_B \in \Delta_{[1:t]}$ always has a lower dimension $t$ than the dimension  $k$ for $p_W$, the inductive procedure will finally terminate.
Hence the estimate $\hat{\bm{p}}_B$ can be obtained, as well as $\hat{\bm{p}}_W$.

The error of each one-step procedure is bounded by the following lemma, proved in Appendix~\ref{sec:pfsuccessive}.

\begin{lemma}
\label{lem:errorreduction}
For $p \geq 2$, we have
\begin{equation}
\label{eq:errorreduction}
\mathbb{E}[\Vert\hat{\bm{p}}_W-\bm{p}_W\Vert_p^p] \leq 2^{p-1}\left(\mathbb{E}[\Vert\hat{\bm{p}}_B-\bm{p}_B\Vert_p^p] + \sum_{s = 1}^t  \mathbb{E}[p_B(s)^{\frac{p}{2}}\hat{p}_B(s)^{\frac{p}{2}}\Vert\hat{\bm{p}}_s - \bm{p}_p\Vert_p^p]\right).
\end{equation}
\end{lemma}

\begin{remark}
\label{rem:errorreductionTV}
For the TV bound ($p = 1$), it is easy to obtain that (cf. Lemma~3.1 in~\cite{Acharya2021})
\begin{equation}
\label{eq:errorreductionTV}
\mathbb{E}[\Vert\hat{\bm{p}}_W-\bm{p}_W\Vert_{\mathrm{TV}}] \leq \mathbb{E}[\Vert\hat{\bm{p}}_B-\bm{p}_B\Vert_{\mathrm{TV}}] + \sum_{s = 1}^t p_B(s) \mathbb{E}[\Vert\hat{\bm{p}}_s - \bm{p}_s\Vert_{\mathrm{TV}}].
\end{equation}
\end{remark}

Now consider the subroutine for estimating all the $\bm{p}_s$, $s = 1,...,t$ given an estimate $\hat{\bm{p}}_B$ for $\bm{p}_B$.  
By~\eqref{eq:errorreduction}, it is intuitive that the resources for estimating each $\bm{p}_s$ should be based on the multiplicative weight $\hat{p}_B(s)^{\frac{p}{2}}p_B(s)^{\frac{p}{2}}$ of the estimation error $\Vert\hat{\bm{p}}_s - \bm{p}_s\Vert_p^p$.
It turns out that the number of encoders for estimating $\bm{p}_s$ can be proportional to $\hat{p}_B(s)$.
Since the quantity $\hat{p}_B(s)$ can be obtained by the decoder, the allocation of encoders can be based on it by interaction between the decoder and encoders.
Such an allocation plan is in contrast to the estimation problem under the TV loss discussed in Appendix~\ref{sec:tvprotocol}. The difference is characterized by the error bound~\eqref{eq:errorreductionTV}, where the weight is simply $p_B(s)$ and a uniform resource allocation plan among all the $\bm{p}_s$, $s = 1,...,t$ is optimal.

The detailed subroutine is presented in the following subsection. 

\subsection{Successive Refinement Subroutines}
\label{subsec:subroutine}
Suppose that there are $m'$ encoders and each of them observes i.i.d. samples $W^n$.
Fix $l_0 \leq l$ and let  $n_0 = \lfloor \frac{l}{l_0}\rfloor \wedge n $.
Then we design the successive refinement subroutine $\mathrm{ASRSub}(m',n,k,l,l_0,p)$  
as follows. It receives an estimate $\hat{\bm{p}}_B$ of the block distribution $\bm{p}_B$ of dimension $t$, and outputs an estimate $\hat{\bm{p}}_W$ of the original distribution~$\bm{p}_W$. 

\subsubsection{Allocating Frames to Blocks}
Devide the $l$-bit message for each encoder into multiple $l_0$-bit frames. Then each encoder holds at least $n_0$ such frames and all encoders hold $m'n_0$ frames in total. Each $l_0$-bit frame is sufficient to transmit a sample, given that the sample is from a fixed block $s$ of size no more than $2^l-1$.
Compute 
\begin{equation}
\label{eq:defrs}
    r(s) = \hat{p}_{B}(s).
\end{equation}
Then $\bm{r}$ is a block distribution.
And all $m'n_0$ frames held by $m'$ encoders we can be allocated for encoding samples in different  $\mathcal{W}_s$, such that 
\begin{enumerate}[(i)]
    \item for each block $s$,  $N_s= \lfloor m'n_0 r(s) \rfloor$  frames are allocated; 
    \item for each encoder, there are at most $\lceil n_0r(s) \rceil$ frames allocated to transmitting samples in $\mathcal{W}_s$.
\end{enumerate}

\subsubsection{Encoding}
For each block $s$, each encoder divides all its $n$ samples into $\lceil n_0r(s) \rceil$ parts, and each part has $\lfloor \frac{n}{\lceil n_0r(s) \rceil} \rfloor$ samples (ignoring the remaining $n-\lceil n_0r(s) \rceil\cdot\lfloor \frac{n}{\lceil n_0r(s) \rceil} \rfloor$).
Each frame that is held by the encoder and allocated for transmitting samples in block $s$ is then mapped to a one of these parts injectively.
If in that part, there are samples falling into the block $s$, then the encoder uses the corresponding frame to encode the first such sample. If not, the frame is encoded as $0$.

\subsubsection{Decoding and Estimating}
For each block $s$, the decoder extracts frames in messages which are allocated to the block. For $l = 1,...,N_s$, let $\tilde{W}^{s}_{l} = \emptyset$ if the $l$-th such frame is $0$ and let $\tilde{W}^{s}_{l}$ be the sample encoded by the frame if it is not $0$. The decoder computes $N_s' = \sum_{l = 1}^{N_s}\mathds{1}_{\tilde{W}^{s}_{l} \neq \emptyset}$. Then it computes
\begin{equation}
    \hat{p}_s(w) = \frac{\sum_{l = 1}^{N_s}\mathds{1}_{\tilde{W}^{s}_{l} = t}}{N_s'}
\end{equation}
if $N_s' \neq 0$, and it computes $\hat{p}_s(w) = \frac{1}{|\mathcal{W}_s|}$ otherwise. 
Finally, for each $s = 1,...,t$ and each $w \in \mathcal{W}_s$, it computes $\hat{p}_W(w) = \hat{p}_B(s) \hat{p}_s(w)$. 

The estimation error induced by the subroutine $\mathrm{ASRSub}(m',n,k,l,l_0,p)$ is described in the following lemma, proved in Appendix~\ref{sec:pfsuccessive}.

\begin{lemma}
\label{lem:errorsub}
For $p\geq 2$, we have 
\begin{equation}
    \sum_{s = 1}^t  \mathbb{E}[p_B(s)^{\frac{p}{2}}\hat{p}_B(s)^{\frac{p}{2}}\Vert\hat{\bm{p}}_s - \bm{p}_s\Vert_p^p]  = O\left( \frac{ \left(1\vee\frac{t}{n^{\frac{p}{2}}} \right) \cdot \left(\frac{l_0}{l} \vee \frac{1}{n}\right)^{\frac{p}{2}}}{m'^{\frac{p}{2}}}  \right).
\end{equation}
\end{lemma}

\subsection{Construction of the Complete Protocol $\mathrm{ASR}$}

Using the subroutine, the complete protocol $\mathrm{ASR}(m,n,k,l,p)$ for the three cases in~\Cref{lem:secsuccessivemain} can be constructed as follows.
Then the error bounds are derived accordingly from~\Cref{lem:errorreduction,lem:errorsub} in Appendix~\ref{sec:pfsuccessive}.

\subsubsection{The Protocol for Case~\ref{case1}}
\label{subsubsec:base}
Invoke the first step of the protocol $\mathrm{AR}(m,n,k,l \wedge k,p)$ in~\Cref{sec:base} and then output the rough estimate~$\hat{\bm{p}}^1_W$. By the analysis in~\ref{subsec:baseanalysis}, we have $\mathbb{E}[\Vert \hat{\bm{p}}_W-\bm{p}_W \Vert_p^p] =  O\left( \left(\frac{k}{mnl}\right)^{\frac{p}{2}} \vee \frac{1}{(mn)^{\frac{p}{2}}}\right)$.

\subsubsection{The Protocol for Case~\ref{case3}}
\label{subsubsec:case3}
Let $l_0 = \lceil \log (\frac{k}{n}+1) \rceil \leq l$ and divide the set $\mathcal{W}$ into $t= \lceil \frac{k}{2^{l_0}-1} \rceil  \in [\frac{n}{2}, n]$ blocks.

Let the first $\frac{m}{2}$ encoders and the decoder estimate the reduced distribution of dimension $t \leq n$. By the assumptions $m(l \wedge n) > 2000 n \log (mn) \log n$, they can invoke the protocol $\mathrm{ASR}(\frac{m}{2},n,t,l,p)$ in~\Cref{subsubsec:base}.

Then let the second $\frac{m}{2}$ encoders and the decoder invoke the subroutine $\mathrm{ASRSub}(\frac{m}{2},n,k,l,l_0,p)$ and compute the estimate of the original distribution $\bm{p}_W$.

\subsubsection{The Protocol for Case~\ref{case4}}
\label{subsubsec:case4}

It suffices to design the protocol for $m \geq \frac{8k}{n2^l}$, since the upper bound is vacuous otherwise.
Let $l_0 = l$ and then compute the integer $a$ as follows. Let $k_1 = k$, then iteratively compute $k_{u+1} = \lceil \frac{k_u}{2^{l}-1} \rceil$ for $u = 1,...,a$. 
Let $a$ be the minimal number satisfying $k_{a+1} \leq n \cdot (2^l-1)$, then $k_{a+1} > n$.

Let the first $\frac{m}{2}$ encoders invoke the protocol $\mathrm{ASR}(\frac{m}{2},n,k,l,p)$  defined in~\Cref{subsubsec:case3} to estimate the last reduced block distribution of dimension $k_{a+1}$.

Divide the second $\frac{m}{2}$ encoders into $a$ parts, such that the $u$-th part has $m_u =  \lfloor \frac{m}{2^{u+1}}\rfloor$ encoders. 
By the choice of $a$, we have $a \leq \left \lceil\frac{2\log(\frac{k}{n(2^l-1)})}{l}\right \rceil$. Then we have $2^{a} \leq 2 \left(\frac{k}{n(2^l-1)}\right)^{\frac{2}{l}} \leq \frac{m}{2}$ for $l \geq 4$, Hence $m_u \geq \frac{m}{2^{a+1}} \geq 1$.
For $u = 1,...,a$, the decoder iteratively invokes $\mathrm{ASRSub}(m_u, n, k_u, l,l_0,p)$ with encoders in the $u$-th part successively.
Then compute the estimate of the original distribution $\bm{p}_W$.

%% file: 6.tex
\section{The Protocol for the Case~\eqref{eq:upperboundgreater2}}
\label{sec:multiplesample}

In this section, we design an adaptive refinement protocol with sample compression that achieves the optimal rates for the case~\eqref{eq:upperboundgreater2}, summarized in the following proposition. 

\begin{proposition}
\label{lem:secmultiplesamplemain}
Let $p \geq 2$, $k > n$, $ml \geq 1000 n \log (mn) \log k$ and $\lceil \log k \rceil \leq l \leq  n^{\frac{2}{p}}$.
Then for the problem in~\Cref{sec:formulation}, there exists an interactive protocol such that for any $\bm{p}_W \in \Delta_{\mathcal{W}}$, the protocol outputs an estimate $\hat{\bm{p}}_W$ satisfying $\mathbb{E}[\Vert \hat{\bm{p}}_W-\bm{p}_W \Vert_p^p] = O\left(\frac{ \log^{\frac{p}{2}} k}{(ml)^{\frac{p}{2}}n^{\frac{p}{2}-1}}\right)$.
\end{proposition}

Note that the communication budget $l \geq \lceil \log k \rceil$ is sufficient to encode more than one samples. 
A naive idea is to let each terminal transmit their i.i.d. samples directly, so that the decoder can infer the distribution based on the samples. 

To achieve higher accuracy, a subset $\mathcal{W}'$ containing $w$ with relatively 
larger $p_W(w)$ is identified and those $p_W(w)$ needs to be refined.
A Sample compression method projects each sample to the subset $\mathcal{W}'$, which makes the encoding of the samples efficient.
The protocol designed in~\Cref{sec:base} is then used to refine the distribution on~$\mathcal{W}'$.
We present the details as follows.

\subsection{The Adaptive Refinement Protocol with Sample Compression}

\subsubsection{Transmit Multiple Samples}
\label{subsubsec:multisample}

Let $n_0 = \lfloor \frac{l}{\lceil \log k \rceil} \rfloor \leq n$. 
Each of the first $\frac{m}{3}$ encoders divides its $l$-bit message into $n_0$ frames, and each frame has $\lceil \log k \rceil$ bits. Then encode each of its first $n_0$ samples by one of these $n_0$ frames. Send the message to the decoder.

Receiving the message, the decoder can access $M_1 \triangleq mn_0$ i.i.d. random samples $(W^1_l)_{l = 1}^{M_1}$.
Then for each $w \in \mathcal{W}$, let
\begin{equation*}
    \hat{\bm{p}}_W^1(w) = \frac{\sum_{l = 1}^{M_1} \mathds{1}_{W^1_l = w}}{M_1}
\end{equation*}
and output the estimate $\hat{\bm{p}}_W^1$.

\subsubsection{Adaptive Refinement with Sample Compression}
\label{subsubsec:shoringup}

Based on the estimate~$\hat{\bm{p}}_W^1$, the decoder computes  
\begin{align*}
&\mathcal{W}' = \left\{w \in \mathcal{W}: \hat{p}^1_W(w) > \frac{2}{n} \right\}, 
\end{align*}
where it is immediate that $|\mathcal{W}'| \leq n-1$ since $\hat{\bm{p}}^1_W$ is normalized.
All the remaining $\frac{2m}{3}$ encoders are informed of $\mathcal{W}'$.

Let the second $\frac{m}{3}$ encoders and the decoder repeat the protocol in~\Cref{subsubsec:multisample}, so that an estimate $\hat{\bm{p}}^2_W(w)$ is obtained by the decoder.

Finally, consider the last $\frac{m}{3}$ encoders.
For the $i$-th encoder among them, it computes $W'_{ij} = h(W_{ij})$ for $j = 1,...,n$, where $(W_{ij})_{j = 1}^n$ are its observed samples and 
\begin{align*}
    h(w) = \left\{\begin{aligned} 
    &w, w \in \mathcal{W}',
    \\
    &\emptyset, w \notin \mathcal{W}'.
    \end{aligned}\right.
\end{align*}
Let $W' = h(W)$ and $p_{W'}$ be its distribution of dimension no more than $n$.
Then each encoder holds $n$ i.i.d. samples $(W'_{ij})_{j = 1}^n$ and $W'_{ij} \sim p_{W'}$.
Let these encoders and the decoder invoke the protocol $\mathrm{AR}(\frac{m}{2},n,|\mathcal{W}'|+1,l,p)$ defined in~\Cref{sec:base} (which is possible since $|\mathcal{W}'|+1 \leq n$ and $ml \geq 1000(|\mathcal{W}'|+1) \log (mn) \log n$). The decoder can obtain the estimate $\hat{\bm{p}}_{W'}^3$ for $\bm{p}_{W'}$. 

Finally, for each $w \in \mathcal{W}$, the decoder computes 
\begin{align*}
    \hat{p}^3_W(w) = \left\{ \begin{aligned}
        &\hat{p}^3_{W'}(w), &w \in \mathcal{W}',
        \\
        &\hat{p}^2_W(w), &w \notin \mathcal{W}',
    \end{aligned}\right.
\end{align*}
and outputs the estimate $\hat{\bm{p}}_W^3$.

\subsection{Error Analysis}
\label{subsec:multiplesampleanalysis}

It is easy to analyze the error for the rough estimate $\hat{\bm{p}}_W^1$. 
For each $w \in \mathcal{W}$, it is folklore that for $p\geq 1$,
\begin{equation}
\label{eq:step1error}
\mathbb{E}[|\hat{p}^1_W(w)-p_W(w)|^p] = O \left( \left(\frac{p_s(w)(1-p_s(w))}{M_1}\right)^{\frac{p}{2}}\right) =  O \left( \left(\frac{p_W(w)}{mn_0}\right)^{\frac{p}{2}}\right).
\end{equation}
For $1 \leq p \leq 2$, taking the summation and using the H\"{o}lder's Inequality imply that
\begin{align*}
\mathbb{E}[\Vert\hat{\bm{p}}^1_W-\bm{p}_W\Vert_p^p] = O \left( \sum_{w \in \mathcal{W}}\frac{p_s(w)^{\frac{p}{2}}}{M_1^{\frac{p}{2}}}\right) \leq O\left(\frac{k^{1-\frac{p}{2}}}{(mn_0)^{\frac{p}{2}}}\right) = O\left(\frac{k^{1-\frac{p}{2}}\log^{\frac{p}{2}} k}{(ml)^{\frac{p}{2}}}\right).  
\end{align*}

The bound is tight up to logarithm factors for $1 \leq p \leq 2$. However, for $p > 2$ we can only get the total error bound $O\left(\frac{\log^{\frac{p}{2}} k}{(ml)^{\frac{p}{2}}}\right)$, which is not tight.
In contrast, the refined estimate $\hat{\bm{p}}_W^3$ can achieve a better upper bound and we show  $\mathbb{E}[\Vert \hat{\bm{p}}^3_W-\bm{p}_W \Vert_p^p] = O\left(\frac{ \log^{\frac{p}{2}} k}{(ml)^{\frac{p}{2}}n^{\frac{p}{2}-1}}\right)$ in Appendix~\ref{sec:pfmultiplesample}.

%% file: 7.tex
\section{The Protocol for Cases~\eqref{eq:upperboundless4} and~\eqref{eq:upperboundgreater4}}
\label{sec:tightbudget}

In this section, we design an adaptive refinement protocol with thresholding that achieves the optimal
rates for cases~\eqref{eq:upperboundless4} and~\eqref{eq:upperboundgreater4}.
It suffices to prove the following proposition in this section. 

\begin{proposition}
\label{lem:sectightbudgetmain}
For the problem in~\Cref{sec:formulation} and each of the following cases, there exists an interactive protocol such that for any $\bm{p}_W \in \Delta_{\mathcal{W}}$, the protocol outputs an estimate $\hat{\bm{p}}_W$ satisfying 
\begin{enumerate}[1.]
    \item \label{caseless} If $1 \leq p \leq 2$, $\lceil \log k \rceil \leq l \leq n$ and $ml<k$, then $\mathbb{E}[\Vert \hat{\bm{p}}_W-\bm{p}_W \Vert_p^p] = O\left(\frac{ \log^{\frac{p}{2}} k}{(ml)^{p-1}}\right)$.
    \item \label{casegreater} If $p>2$, $\lceil \log k \rceil \leq l \leq n$ and $ml < n$, then $\mathbb{E}[\Vert \hat{\bm{p}}_W-\bm{p}_W \Vert_p^p] = O\left(\frac{ \log^{p} k \vee \log^{p} (mn) \log^{p} n}{(ml)^{p-1}} \vee \frac{1}{(mn)^{\frac{p}{2}}}\right)$.
\end{enumerate}

\end{proposition}

To overcome the difficulty induced by the extremely tight total communication budget, huge "preys" and little "flies" among all $p_W(w)$ to be estimated should be classified and dealt with differently.
The thresholding level is naturally $~\frac{1}{ml}$, since roughly $\sim ml$ samples can be transmitted by the protocol in~\Cref{subsubsec:multisample}.
For those little "flies" $p_W(w) \preceq \frac{1}{ml}$, it is better to overlooking them than trying to estimating them. 
The remaining resources should be focused on refining huge "preys" $p_W(w) \succeq \frac{1}{ml}$, whose number $\sim ml$ is limited.
For $p > 2$, sample compression methods and the protocol in~\Cref{sec:base} are applied to refine the estimate similar to the protocol in~\Cref{subsubsec:shoringup}.
With the help of thresholding methods, the resulting estimation protocol can catch the rough landscape of the distribution $\bm{p}_W$ and achieve the optimal error rate under the communication constraints.

We present the protocols for two cases respectively in the following subsections and detailed error analysis can be found in~\Cref{sec:pftightbudget}.

\subsection{Thresholding Methods for Case~\ref{caseless}}
\label{subsec:thresholding}

\subsubsection{Rough Estimation}
Let $n_0 = \lfloor \frac{l}{\lceil \log k \rceil} \rfloor \leq n$. Let the first $\frac{m}{2}$ encoders and the decoder invoke the protocol presented in~\Cref{subsubsec:multisample}, so that the decoder can obtain an estimate $\hat{\bm{p}}^1_W$.

\subsubsection{Thresholding Step}
Based on that, the decoder computes  
\begin{align*}
&\mathcal{W}' = \left\{w \in \mathcal{W}: \hat{p}^1_W(w) > \frac{2}{ml} \right\}, 
\end{align*}
where it is immediate that $|\mathcal{W}'| \leq ml$ since $\hat{\bm{p}}^1_W$ is normalized.

Let the second $\frac{m}{2}$ encoders and the decoder repeat the protocol in~\Cref{subsubsec:multisample}, so that an estimate $\hat{\bm{p}}^2_W(w)$ is obtained by the decoder.

Then for each $w \in \mathcal{W}$, the decoder computes 
\begin{align*}
    \hat{p}^3_W(w) = \left\{ \begin{aligned}
        &\hat{p}^2_{W}(w), &w \in \mathcal{W}',
        \\
        &0, &w \notin \mathcal{W}',
    \end{aligned}\right.
\end{align*}
and outputs the estimate $\hat{\bm{p}}_W^3$.

\subsection{Combining Thresholding Methods and Refinement for Case~\ref{casegreater}}
\label{subsec:combining}

\subsubsection{Rough Estimation}
Let $k' = \frac{ml}{2000\log (mn) \log n}$, then $k' < ml < n$ and $ml > 1000 k' \log (mn) \log n$. 

Let the first $\frac{m}{2}$ encoders and the decoder invoke the protocol presented in~\Cref{subsubsec:multisample}.
Then the decoder can obtain an estimate $\hat{\bm{p}}^1_W$.

\subsubsection{The Mixed Thresholding and Refinement Mechanism}
Based on that, the decoder computes  
\begin{align*}
&\mathcal{W}' = \left\{w \in \mathcal{W}: \hat{p}^1_W(w) > \frac{2}{k'} \right\}, 
\end{align*}
where it is immediate that $|\mathcal{W}'| \leq k'-1$ since $\hat{\bm{p}}^1_W$ is normalized.
All the remaining $\frac{m}{2}$ encoders are informed of~$\mathcal{W}'$.

Then consider the second $\frac{m}{2}$ encoders.
For the $i$-th encoder among them, it computes $W'_{ij} = h(W_{ij})$ for $j = 1,...,n$, where $(W_{ij})_{j = 1}^n$ are its observed samples and 
\begin{align*}
    h(w) = \left\{\begin{aligned} 
    &w, w \in \mathcal{W}',
    \\
    &\emptyset, w \notin \mathcal{W}'.
    \end{aligned}\right.
\end{align*}
Let $W' = h(W)$ and $p_{W'}$ be its distribution of dimension no more than $n$.
Then each encoder holds $n$ i.i.d. samples $(W'_{ij})_{j = 1}^n$ and $W'_{ij} \sim p_{W'}$.
Let these encoders and the decoder invoke the protocol $\mathrm{ASR}(\frac{m}{2},n,|\mathcal{W}'|+1,l,p)$ defined in~\Cref{sec:base} (which is possible since $|\mathcal{W}'|+1 \leq k' < n$ and $ml \geq 1000(|\mathcal{W}'|+1) \log (mn) \log n$). The decoder can obtain the estimate $\hat{\bm{p}}_{W'}^2$ for $\bm{p}_{W'}$. Then for each $w \in \mathcal{W}$, it computes 
\begin{align*}
    \hat{p}^3_W(w) = \left\{ \begin{aligned}
        &\hat{p}^2_{W'}(w), &w \in \mathcal{W}',
        \\
        &0, &w \notin \mathcal{W}',
    \end{aligned}\right.
\end{align*}
and outputs the estimate $\hat{\bm{p}}_W^3$.

%% file: 8.tex
\section{The Protocol for $n = 1$}
\label{sec:onesample}

In this section, we design a non-interactive protocol based on random hashing, which achieves the optimal
rate for~$n = 1$.
Similar to the discussion in~\Cref{rem:Holder}, it suffices to show the following proposition for $p \geq 2$.

\begin{proposition}
\label{lem:seconesamplemain}
Let $p \geq 2$ and $n = 1$.
Then there exists a non-interactive protocol such that for any $\bm{p}_W \in \Delta_{\mathcal{W}}$, the protocol outputs an estimate $\hat{\bm{p}}_W$ satisfying $\mathbb{E}[\Vert \hat{\bm{p}}_W-\bm{p}_W \Vert_p^p] = O\left(\frac{k}{(m2^l)^{\frac{p}{2}}} \vee \frac{1}{m^{\frac{p}{2}}}\right)$.
\end{proposition}

\subsection{Motivation of the Protocol}

The most natural idea is to first invoke the simulation protocol in~\cite{Acharya2020II} to output $M = O(\frac{m2^l}{k})$ samples from the distribution $\bm{p}_{\mathcal{W}}$ at the decoder side; then estimate $\bm{p}_{\mathcal{W}}$ using $M$ samples by a traditional central estimation method.
It can achieve the optimal minimax rate $\frac{k}{m2^l}$ for $p=2$, and hence the optimal rate $\frac{k}{(m2^l)^{\frac{p}{2}}}$ for $1 \leq p \leq 2$.
However, for $p \geq 2$, using $M$ i.i.d. samples to estimate the underlying distribution under the $\ell^p$ loss can only achieve a rate of $\frac{1}{M^{\frac{p}{2}}} = (\frac{k}{m2^l})^{\frac{p}{2}}$ , which leaves a gap with the lower bound $\frac{k}{(m2^l)^{\frac{p}{2}}}$ by~\Cref{lem:lowerboundless}.
The above naive protocol is not optimal and we can show that the lower bound $\frac{k}{(m2^l)^{\frac{p}{2}}}$ is optimal.

The subtle difference is that the minimax optimal rate without the communication constraint is $\frac{1}{M^{\frac{p}{2}}}$ for $p \geq 2$ (cf.~\Cref{lem:lowerboundcentral}), in contrast with the optimal rate $\frac{k^{1-\frac{p}{2}}}{M^{\frac{p}{2}}}$ for $1 \leq p \leq 2$. The difference was ignored by the proof of upper bound in some previous work~\cite{Chen2024}, hence the optimal rate claimed therein is not true. Constructing the order-optimal protocol really deserves special care, which is the main goal in the remaining part of this section.

The aforementioned difficulty in estimation under $\ell^p$ losses can be overcome, by using a random hash function to compress the sample first, and then constructing and rescaling the histogram to obtain the estimate.
No simulation step as in~\cite{Acharya2020II} is needed. 
Moreover, it is worth mentioning that the resulting protocol is non-interactive. 
The idea is similar to the second estimation stage in~\cite{Chen2021} for estimating a sparse distribution under communication constraints.
Details of the protocol are presented as follows, and the error analysis can be found in Appendix~\ref{sec:pfonesample}.

\subsection{The Non-interactive Protocol Based on Random Hashing for $n = 1$}

\subsubsection{Encoding}
Let the $i$-th encoder generate a random hash function $h_{i}: \mathcal{W} \to \{0,1\}^l$, $i = 1,...,m$ by shared randomness (i.e. $(h_i(w))_{w \in \mathcal{W}}$ are independent and $\mathbb{P}[h_i(w) = b] = 2^{-l}$ for each $w \in \mathcal{W}$ and $b \in \{0,1\}^l$), so that the decoder can also generate $h_i$.
Observing its sample~$W_i$, the $i$-th encoder computes $B_i = h_i(W_i)$ and sends it to the decoder.

\subsubsection{Decoding}
Upon receiving $B_i$, the decoder then computes
\begin{equation}
\label{eq:rescaling}
    \hat{p}_W(w) = \frac{2^l}{2^l-1} \cdot \frac{\sum_{i = 1}^{m} \mathds{1}_{h_i(w) = B_i}}{m}-\frac{1}{2^l-1}
\end{equation}
for each $w \in \mathcal{W}$ and outputs $\hat{\bm{p}}_W$. 

%% file: 9.tex
\section{Lower Bounds}
\label{sec:lowerbound}

In order to prove~\Cref{lem:lowerboundless,lem:lowerboundgreater}, we first reorganize the lower bounds into the following three lemmas.

\begin{lemma}
\label{lem:lowerboundregular}
   For $1 \leq p \leq 2$, we have
\begin{equation*}
    R(m,n,k,l,p) \succeq \left\{
    \begin{aligned}
        &\frac{k}{(mnl)^{\frac{p}{2}}}, &n \geq k \log k,\  &m > \left(\frac{k}{l}\right)^2,\  l \leq k,\\
        &\frac{k^{1-\frac{p}{2}}}{(ml\log k)^{\frac{p}{2}}}, &n < k \log k,\  &m > \left(\frac{k}{l}\right)^2,\  l \leq \frac{n}{\log k}, \\
        &\frac{k}{(mn2^l)^{\frac{p}{2}}},  &  &mn2^l > k^2.
    \end{aligned}\right.
\end{equation*}
For $p \geq 2$, we have
\begin{equation*}
    R(m,n,k,l,p) \succeq \left\{
    \begin{aligned}
        &\frac{k}{(mnl)^{\frac{p}{2}}}, &n \geq k \log k,\  &m > \left(\frac{k}{l}\right)^2,\  l \leq k^{\frac{2}{p}},\\
        &\frac{1}{(ml)^{\frac{p}{2}}n^{\frac{p}{2}-1}\log n}, &n < k \log k,\  &m > \left(\frac{n/\log n}{l}\right)^2, \ l \leq \left(\frac{n}{\log n}\right)^{\frac{2}{p}}, \\
        &\frac{k}{(mn2^l)^{\frac{p}{2}}},  & &mn2^l > k^2.
    \end{aligned}\right.
\end{equation*}
\end{lemma}

\begin{lemma}
\label{lem:lowerboundcentral}
For $1 \leq p \leq 2$, $R(m,n,k,l,p) \succeq \frac{k^{1-\frac{p}{2}}}{(mn)^{\frac{p}{2}}}$.
For $p \geq 2$, then $R(m,n,k,l,p) \succeq \frac{1}{(mn)^{\frac{p}{2}}}$.
\end{lemma}

\begin{lemma}
\label{lem:lowerboundtightbudget}
If $2ml < k$, then $R(m,n,k,l,p) \succeq \frac{1}{(ml)^{p-1}}$.
\end{lemma}

\Cref{lem:lowerboundregular} is proved by exploiting the results for $p = 1$ in~\cite{Acharya2021}, and details can be found in Appendix~\ref{pflem:lowerboundregular}.
We show~\Cref{lem:lowerboundcentral,lem:lowerboundtightbudget} in~\Cref{pflem:lowerboundcentral,pflem:lowerboundtightbudget}, respectively.


\subsection{Proof of Lemma~\ref{lem:lowerboundcentral}}
\label{pflem:lowerboundcentral}
The results for $1 \leq p \leq 2$ are well-known~\cite{Acharya2023a,Chen2024}, hence we only give the proof for $p \geq 2$. We use the information-theoretic methods.

\subsubsection{Choose a  prior distribution and lower bound the minimax risk by the Bayes risk} 
We can assume that $\mathcal{W} = [1:k]$ without loss of generality. Let 
\begin{equation}
    \begin{aligned}
        p^1_{W} = \left(\frac{1+\epsilon}{2}, \frac{1-\epsilon}{2},0,...,0\right),
        \\
        p^2_{W} = \left(\frac{1-\epsilon}{2}, \frac{1+\epsilon}{2},0,...,0\right).
    \end{aligned}
\end{equation}
Let $Z \sim \mathrm{Bern}(\frac{1}{2})$ and define the prior distribution to be $p^Z_{W}$. 
Let $\mathcal{P}$ be an $(m,n,l)$-protocol defined in~\Cref{sec:formulation}, 
then we have
\begin{align*}
    \sup_{p_{W} \in \Delta_{\mathcal{W}}} \mathbb{E}[\Vert\hat{p}_{W}^{\mathcal{P}}-p_{W}\Vert_p^p] 
    \geq & \frac{1}{2} \left( \mathbb{E}[\Vert\hat{p}_{W}^{\mathcal{P}}-p_{W}^1 \Vert_p^p] + \mathbb{E}[\Vert\hat{p}_{W}^{\mathcal{P}}-p_{W}^2 \Vert_p^p] \right)
    \\
    = & \mathbb{E}[\Vert\hat{p}_{W}^{\mathcal{P}}-p^Z_{W}\Vert_p^p].
\end{align*}

\subsubsection{Convert the estimation problem into a testing problem}
Let 
\begin{equation*}
    \hat{Z} = \mathop{\arg\min}_{z \in \{0,1\}} \Vert p_{W}^{z}-\hat{p}_{W}^{\mathcal{P}}\Vert_p.
\end{equation*}
Then we have 
\begin{align*}
    \Vert p_{W}^{\hat{Z}}-p^Z_{W}\Vert_p \leq & \Vert\hat{p}_{W}^{\mathcal{P}}-p_{W}^{\hat{Z}}\Vert_p + \Vert\hat{p}_{W}^{\mathcal{P}}-p^Z_{W}\Vert_p  
    \\
    \leq &2 \Vert\hat{p}_{W}^{\mathcal{P}}-p^Z_{W}\Vert_p.
\end{align*}
Hence we have 
\begin{equation}
\label{eq:lowerbyaverage}
\begin{aligned}
\mathbb{E}[\Vert\hat{p}_{W}^{\mathcal{P}}-p^Z_{W}\Vert^p_p] \geq &\frac{1}{2^p}\mathbb{E}[ \Vert p_{W}^{\hat{Z}}-p^Z_{W}\Vert^p_p]
\\
=&\frac{1}{2^{p-1}}\epsilon^p \mathbb{P}[\hat{Z} \neq Z].
\end{aligned}
\end{equation}

Since $Z-W^{mn}-B^m-\hat{Z}$ is a Markov chain, then by the Fano's inequality, we have
\begin{equation}
\label{eq:generalizedFano}
    I(Z;B^m) \geq 1- h\left(\mathbb{P}[\hat{Z} \neq Z]\right),
\end{equation}
where $h(p) = -p \log_2 p - (1-p) \log_2 (1-p)$ is the binary entropy function.
If we can show that for a suitably chosen $\epsilon$,  
\begin{equation}
\label{eq:mutualupper}
I(Z;B^m) \leq \frac{1}{2},
\end{equation}
then by~\eqref{eq:lowerbyaverage} and~\eqref{eq:generalizedFano} we have 
$$
 \mathbb{P}[\hat{Z} \neq Z]  \geq \frac{1}{10k},
$$
thus
$$
\mathbb{E}[\Vert\hat{p}_{W}^{\mathcal{P}}-p_{W}^{Z}\Vert^2_2] \succeq \epsilon^p.
$$
Then we have $R(m,n,l,r) \succeq \epsilon^p$.

\subsubsection{Choose a suitable parameter}
By the Markov chain $Z_s-W^{mn}-B^m$ and the data processing inequality, we have
\begin{align*}
    &I(Z;B^m) \leq I(Z;W^{mn})
   \\
    = & \frac{1}{2} D_{\mathcal{W}^{mn}}\left(p_{W}^{1}(w^{mn}) ||\frac{1}{2} \left( p_{W}^{1}(w^{mn})+p_{W}^{2}(w^{mn})\right) \right) + \frac{1}{2} D_{\mathcal{W}^{mn}}\left(p_{W}^{2}(w^{mn}) ||\frac{1}{2} \left( p_{W}^{1}(w^{mn})+p_{W}^{2}(w^{mn})\right) \right)
    \\
    \leq & \frac{1}{4}  \left(D_{\mathcal{W}^{mn}}\left(p^1_{W}(w^{mn}) || p^2_{W}(w^{mn})\right) + D_{\mathcal{W}^{mn}}\left(p^2_{W}(w^{mn}) || p^1_{W}(w^{mn})\right)\right)
    \\
    = & \frac{mn}{2} D_{\mathcal{W}}\left(p^2_{W}(w) || p^1_{W}(w)\right)
    \\
    = & \frac{mn\epsilon}{2} \log \left( 1+\frac{2\epsilon}{1-\epsilon}\right)
    \\
    \leq & \frac{mn\epsilon^2}{1-\epsilon}, 
\end{align*}
where the first inequality is due to the convexity of KL divergence and the second is by the fact that $\log(1+x) \leq x$ for $x>0$. 
By letting $\epsilon =(100mn)^{-\frac{1}{2}}$ we obtain that $R(m,n,l,r) \succeq (mn)^{-\frac{p}{2}}$.

\subsection{Proof of Lemma~\ref{lem:lowerboundtightbudget}}
\label{pflem:lowerboundtightbudget}

The case for $ml < k$ is not hard, but it has not been fully explored in previous literature.
First note that by the H\"{o}lder's inequality, we have
\begin{equation*}
    \Vert\hat{\bm{p}}_W-\bm{p}_W\Vert_{\mathrm{TV}} \leq k^{1-\frac{1}{p}} \Vert\hat{\bm{p}}_W-\bm{p}_W\Vert_p.
\end{equation*}
Hence we have 
\begin{equation}
\label{eq:1top}
    R(m,n,k,l,p) \geq k^{1-p} R(m,n,k,l,1)^{p},
\end{equation}
and the minimax lower bound for $p \geq 1$ is easily implied by that for $p = 1$.

We have the following folklore lemma for $p = 1$, which can be proved by the Fano's method and the data processing inequality.

\begin{lemma}
\label{lem:lowerbound''}
 If $2ml \leq k$, then we have $R(m,n,k,l,1) \succeq 1$.
\end{lemma}

Combining~\Cref{lem:lowerbound''} and~\eqref{eq:1top}, for any $k\geq 2ml$ we have
\begin{equation*}
    R(m,n,k,l,p)  \succeq \frac{1}{k^{p-1}}.
\end{equation*}
Hence we further have 
\begin{equation*}
    R(m,n,k,l,p) \geq R(m,n,2ml,l,p) \succeq \frac{1}{(ml)^{p-1}}.
\end{equation*}

%% file: 10.tex
\section{Discussions}
\label{sec:discussion}

Note that the methods in this work are not restricted to the discrete distribution estimation problem.
The analysis of statistical learning problems in various other settings under $\ell^p$ losses can also benefit from our methods.  
The methods deal with the difficulty induced by the normalization constraint of the distribution in the distribution estimation setting, which also shows a potential direction for solving problems with similar implicit constraints. 
A more challenging problem is whether we can construct non-interactive protocols, instead of interactive protocols in this work, to achieve the minimax optimal rates with $n>1$ samples per terminal and under $\ell^p$ losses. 
Determining the privacy-constrained optimal rates for $n>1$ and $\ell^p$ losses is also an interesting direction for future work.


%% file: A1.tex
\appendices
\section{Proof of Proposition~\ref{lem:secbasemain}: Error Analysis for the Protocol in Section~\ref{sec:base}}
\label{sec:pfbase}

We first show the following preliminary error bound concerning the rough estimate.

\begin{lemma}
\label{lem:crudeerrorprobability'}
If $p_W(w) \geq \frac{1}{k}$ for some $w \in \mathcal{W}$, then $\mathbb{P}\left[\frac{p_W(w)}{\hat{p}^1_W(w)}\geq 2 \right] \leq O\left(\frac{1}{(np_W(w))^{\frac{p}{2}}}\right)$.
\end{lemma}

\begin{IEEEproof}
By~\eqref{eq:crudeerror} and $p_W(w) \geq \frac{1}{k}$, we have
\begin{equation}
\label{eq:crudeerror2}
\begin{aligned}
\mathbb{E}\left[|p_W(w)-\hat{p}^1_W(w)|^p\right] 
= O\left( \left(\frac{kp_W(w)}{mnl}\right)^{\frac{p}{2}}\right).
\end{aligned}
\end{equation}

By the Markov inequality, we can obtain that
\begin{align*}
    \mathbb{P}\left[\frac{p_W(w)}{\hat{p}^1_W(w)}\geq 2 \right] = \mathbb{P}\left[\frac{\hat{p}^1_W(w)}{p_W(w)}\leq \frac{1}{2} \right] \leq \mathbb{P}\left[ \left|\hat{p}^1_W(w)-p_W(w)\right|\geq \frac{1}{2} p_W(w)\right] \leq \frac{2^{p}\mathbb{E}[\left|\hat{p}^1_W(w)-p_W(w)\right|^p]}{p_W(w)^p}.
\end{align*}

Then by~\eqref{eq:crudeerror2} and the assumption that $ml >1000 k \log (mn) \log n$, we complete the proof.
\end{IEEEproof}

Now we return to the proof of~\Cref{lem:secbasemain}.
Note that it suffices to show that for each $w \in \mathcal{W}$, 
\begin{equation}
\label{eq:baseerror}
    \mathbb{E}\left[|p_W(w)-\hat{p}^2_W(w)|^p\right]  = O\left(\frac{1}{(mnl)^{\frac{p}{2}}} + \frac{p_W(w)^{\frac{p}{2}}}{(mn)^{\frac{p}{2}}}\right),
\end{equation}
then taking the summation can complete the proof. 

By~\Cref{lem:binaryestimate}, we have
\begin{align*}
    &\mathbb{E}\left[|p_W(w)-\hat{p}^2_W(w)|^p\right] = O\left( \mathbb{E}\left[\left(\frac{p_W(w)}{mnl(\hat{p}^1_W(w)+\frac{1}{k})}\right)^{\frac{p}{2}}\right] + \left(\frac{1}{mnl}\right)^{\frac{p}{2}} + \left(\frac{p_W(w)}{mn}\right)^{\frac{p}{2}} \right).
\end{align*}
It suffices to bound the first term.
Define the event $\mathcal{F}_w = \left\{\frac{p_W(w)}{\hat{p}^1_W(w)}\geq 2\right\}$. Then by~\Cref{lem:crudeerrorprobability'} and $n \geq k$, we have
\begin{align*}
    &\mathbb{E}\left[\left(\frac{p_W(w)}{mnl(\hat{p}^1_W(w)+\frac{1}{k})}\right)^{\frac{p}{2}}\right] 
    \\
    = &\mathbb{E}\left[ \mathds{1}_{\mathcal{F}_w}\left(\frac{p_W(w)}{mnl(\hat{p}^1_W(w)+\frac{1}{k})}\right)^{\frac{p}{2}}\right] + \mathbb{E}\left[ \mathds{1}_{\mathcal{F}_w^{\complement}}\left(\frac{p_W(w)}{mnl(\hat{p}^1_W(w)+\frac{1}{k})}\right)^{\frac{p}{2}}\right] 
    \\
    \leq & \mathbb{P}\left[\mathcal{F}_w\right] \cdot \left(\frac{kp_W(w)}{mnl}\right)^{\frac{p}{2}}+ O\left( \left(\frac{1}{mnl}\right)^{\frac{p}{2}} \right)  
    \\
    = &\mathds{1}_{\{p_W(w) < \frac{1}{k}\}} \cdot O\left( \left(\frac{1}{mnl}\right)^{\frac{p}{2}} \right) + \mathds{1}_{\{p_W(w) \geq \frac{1}{k}\}} \cdot  O\left(\left(\frac{1}{np_W(w)} \cdot \frac{kp_W(w)}{mnl}\right)^{\frac{p}{2}}\right) + O\left( \left(\frac{1}{mnl}\right)^{\frac{p}{2}} \right)
    \\
    = & O\left( \left(\frac{1}{mnl}\right)^{\frac{p}{2}} \right),
\end{align*}
which completes the proof.

%% file: A2.tex
\section{Proof of Proposition~\ref{lem:secsuccessivemain}: Error Analysis for the protocol in Section~\ref{sec:successive}}
\label{sec:pfsuccessive}

\subsection{Proof of Lemma~\ref{lem:errorreduction}}

Note that
\begin{align*}
    &(p_B(s) p_s(w) - \hat{p}_B(s) \hat{p}_s(w))^2
    \\
    \leq &(p_B(s) p_s(w) - \hat{p}_B(s) \hat{p}_s(w))^2 + (p_B(s) \hat{p}_s(w) - \hat{p}_B(s) p_s(w))^2
    \\
    = &(p_s(w)^2+ \hat{p}_s(w)^2) (p_B(s) -\hat{p}_B(s))^2 + 2p_B(s)\hat{p}_B(s)( p_s(w) -  \hat{p}_s(w))^2.
\end{align*}
Then by the H\"{o}lder's inequality, we have 
\begin{align*}
    &(p_B(s) p_s(w) - \hat{p}_B(s) \hat{p}_s(w))^p
    \\
    \leq & 2^{\frac{p}{2}-1}\left[\left(p_s(w)^2+ \hat{p}_s(w)^2\right)^{\frac{p}{2}} (p_B(s) -\hat{p}_B(s))^p + 2^{\frac{p}{2}}p_B(s)^{\frac{p}{2}}\hat{p}_B(s)^{\frac{p}{2}}( p_s(w) -  \hat{p}_s(w))^p\right]
    \\
    \leq & 2^{p-1}\left[\frac{1}{2}\left(p_s(w)+ \hat{p}_s(w)\right) (p_B(s) -\hat{p}_B(s))^p + p_B(s)^{\frac{p}{2}}\hat{p}_B(s)^{\frac{p}{2}}( p_s(w) -  \hat{p}_s(w))^p\right].
\end{align*}
where the last inequality is since $p \geq 2$ and $p_s(w), \hat{p}_s(w) \in [0,1]$. 
Take the summation, and then we have 
\begin{align*}
    \Vert \hat{\bm{p}}_W - \bm{p}_W \Vert_p^p \leq &2^{p-1}\sum_{s = 1}^t \left[ (p_B(s) -\hat{p}_B(s))^p + p_B(s)^{\frac{p}{2}}\hat{p}_B(s)^{\frac{p}{2}} \Vert \hat{\bm{p}}_s-\bm{p}_s \Vert^p \right].
\end{align*}
Then~\eqref{eq:errorreduction} is obtained by taking the expectation. We complete the proof.

\subsection{Proof of Lemma~\ref{lem:errorsub}}

If $m'n_0r(s) = m'n_0\hat{p}_B(s) \leq 4$, since $\Vert\hat{\bm{p}}_s - \bm{p}_s\Vert_p^p \leq 2$, then
$$
p_B(s)^{\frac{p}{2}}\hat{p}_B(s)^{\frac{p}{2}}\Vert\hat{\bm{p}}_s - \bm{p}_s\Vert_p^p \leq 2p_B(s)^{\frac{p}{2}}\hat{p}_B(s)^{\frac{p}{2}} \leq 2^{p+1}\left( \frac{p_B(s)}{m'n_0}\right)^{\frac{p}{2}}.
$$

Otherwise, we have $m'n_0r(s) = m'n_0\hat{p}_B(s) > 4$, hence $N_s = \Theta \left( m'n_0r(s) \right) = \Theta \left( m'n_0\hat{p}_B(s)\right)$.
Given $\hat{\bm{p}}_B$, then $\tilde{W}^{s}_{u}$ for $u = 1,...,N_s$ are i.i.d. random variables with
\begin{equation}
    q_s \triangleq \mathbb{P}[\tilde{W}^{s}_{u} \neq \emptyset|\hat{\bm{p}}_B] = 1-(1-p_B(s))^{\lfloor \frac{n}{\lceil n_0 r(s) \rceil} \rfloor} = \Theta \left(p_B(s) \left\lfloor \frac{n}{\lceil n_0 r(s) \rceil} \right\rfloor \wedge 1\right) =  \Theta \left(p_B(s) \left\lfloor \frac{n}{\lceil n_0\hat{p}_B(s) \rceil} \right\rfloor \wedge 1\right). 
\end{equation}
In this case, we can establish the bound shown in the following lemma.

\begin{lemma}
$\mathbb{E}[p_B(s)^{\frac{p}{2}}\hat{p}_B(s)^{\frac{p}{2}}\Vert\hat{\bm{p}}_s - \bm{p}_s\Vert_p^p|\hat{\bm{p}}_B]  \leq C\mathbb{E}\left[ \left( \frac{ \hat{p}_B(s) }{m'n} \vee \frac{1}{m'nn_0} \vee \frac{p_B(s)}{m'n_0}\right)^{\frac{p}{2}}\Big|\hat{\bm{p}}_B\right]$ for some $C>0$.
\end{lemma}

\begin{IEEEproof}
By the Chernoff bound, we have 
\begin{equation}
    \mathbb{P}\left[N_s' \geq \frac{N_s q_s}{2}\Big| \hat{\bm{p}}_B\right] \leq \exp \left(-\frac{N_s q_s}{8} \right).
\end{equation}
And conditional on the event $\{\tilde{W}^{s}_{u} \neq \emptyset \}$, the distribution of $\tilde{W}^{s}_{u}$ is $\bm{p}_s$. Hence for each $w \in \mathcal{W}_s$, it is folklore that (cf. Theorem 4 in~\cite{Skorski2020}),
$$
\mathbb{E}[|\hat{p}_s(w)-p_s(w)|^p|N_s',\hat{\bm{p}}_B] = O \left( \left(\frac{p_s(w)(1-p_s(w))}{N_s'}\right)^{\frac{p}{2}}\right).
$$
Take the summation, since $p\geq 2$ and $p_s(w) \in [0,1]$ we have
\begin{align*}
\mathbb{E}\left[\Vert\hat{\bm{p}}_s - \bm{p}_s\Vert_p^p \Big| N_s' \geq \frac{N_s q_s}{2},\hat{\bm{p}}_B\right] = O\left( \frac{1}{N_s^{\frac{p}{2}} q_s^{\frac{p}{2}}} \right). 
\end{align*}
Since $\Vert\hat{\bm{p}}_s - \bm{p}_s\Vert^2 \leq 2$, we have
\begin{align*}
\mathbb{E}[\Vert\hat{\bm{p}}_s - \bm{p}_s\Vert^2|\hat{\bm{p}}_B] \leq 2\exp \left(-\frac{N_s q_s}{8} \right)+ O\left( \frac{1}{N_s^{\frac{p}{2}} q_s^{\frac{p}{2}}} \right) = O\left( \frac{1}{N_s^{\frac{p}{2}} q_s^{\frac{p}{2}}} \right). 
\end{align*}

Since $n_0 \leq n$, we have $\lceil n_0\hat{p}_B(s)\rceil \leq n$ and $ \frac{n}{\lceil n_0\hat{p}_B(s) \rceil} \geq 1$. Hence there exists some $C>0$, such that
\begin{align*}
    & \mathbb{E}[p_B(s)^{\frac{p}{2}}\hat{p}_B(s)^{\frac{p}{2}}\Vert\hat{\bm{p}}_s - \bm{p}_s\Vert_p^p|\hat{\bm{p}}_B] 
    \\
    \leq &C \mathbb{E}\left[\left( \frac{p_B(s)\hat{p}_B(s)}{m'n_0\hat{p}_B(s) q_s} \right)^{\frac{p}{2}}\Big|\hat{\bm{p}}_B\right]
    \\
    = & C\mathbb{E}\left[ \left( \frac{p_B(s)}{m'n_0 \left(p_B(s) \lfloor \frac{n}{\lceil n_0\hat{p}_B(s) \rceil} \rfloor \wedge 1\right)} \right)^{\frac{p}{2}}\Big|\hat{\bm{p}}_B\right]
    \\
    = &C \mathbb{E}\left[ \left( \frac{1}{m'n_0 \left( \frac{n}{\lceil n_0\hat{p}_B(s) \rceil} \right)} \vee \frac{p_B(s)}{m'n_0} \right)^{\frac{p}{2}}\Big|\hat{\bm{p}}_B\right]
    \\
    = &C \mathbb{E}\left[ \left( \frac{ n_0\hat{p}_B(s) \vee 1}{m'nn_0 } \vee \frac{p_B(s)}{m'n_0}\right)^{\frac{p}{2}}\Big|\hat{\bm{p}}_B\right],
\end{align*}
completing the proof.
\end{IEEEproof}

In both cases, we can take the expectation and obtain that 
\begin{align*}
    \mathbb{E}[p_B(s)^{\frac{p}{2}}\hat{p}_B(s)^{\frac{p}{2}}\Vert\hat{\bm{p}}_s - \bm{p}_s\Vert_p^p] \leq C'\mathbb{E}\left[ \left( \frac{ \hat{p}_B(s) }{m'n} \vee \frac{1}{m'nn_0} \vee \frac{p_B(s)}{m'n_0}\right)^{\frac{p}{2}}\right],
\end{align*}
for some $C'>0$.

Finally, take the sum over $s$ and note that $p \geq 2$, then 
\begin{align*}
    &\sum_{s = 1}^t  \mathbb{E}[p_B(s)^{\frac{p}{2}}\hat{p}_B(s)^{\frac{p}{2}}\Vert\hat{\bm{p}}_s - \bm{p}_s\Vert_p^p]
    \\
    \leq &C' \sum_{s = 1}^t \mathbb{E}\left[\left( \frac{ \hat{p}_B(s) }{m'n} \vee \frac{1}{m'nn_0} \vee \frac{p_B(s)}{m'n_0}\right)^{\frac{p}{2}}\right]
    \\
    = &O\left(\left( \frac{1}{m'n_0}\right)^{\frac{p}{2}} \vee \frac{t}{(m'nn_0)^{\frac{p}{2}}} \right)
    \\
    = &O\left( \frac{ \left(1\vee\frac{t}{n^{\frac{p}{2}}} \right) \cdot \left(\frac{l_0}{l} \vee \frac{1}{n}\right)^{\frac{p}{2}}}{m'^{\frac{p}{2}}}  \right),
\end{align*}
which completes the proof.

\subsection{Proof of Proposition~\ref{lem:secsuccessivemain}: Analysis of The Protocol for Case~\ref{case3}}
\label{subsec:analysiscase3}
By the case~\ref{case1}, the estimation error for the reduced block distribution is bounded by
\begin{align*}
     C_3 \cdot \left[ \left(\frac{t}{mnl}\right)^{\frac{p}{2}} \vee \frac{1}{(mn)^{\frac{p}{2}}} \right]
\end{align*}
for some $C_3>0$.

By~\Cref{lem:errorsub}, the estimation error for the conditional distribution induced by the invoking of the subroutine $\mathrm{ASRSub}(\frac{m}{2},n,k,l,l_0,p)$ is bounded by
\begin{align*}
    C_4 \cdot \left(\frac{\left(\frac{l_0}{l} \vee \frac{1}{n}\right)}{\frac{m}{2}}\right)^{\frac{p}{2}} = C_4 \left(\frac{2\left(\frac{l_0}{l} \vee \frac{1}{n}\right)}{m}\right)^{\frac{p}{2}} = C_4 \left[\left(\frac{2l_0}{ml}\right)^{\frac{p}{2}} \vee \frac{2^{\frac{p}{2}}}{(mn)^{\frac{p}{2}}}\right] \leq  C_4 \left[ \left(\frac{4\log (\frac{k}{n}+1)}{ml}\right)^{\frac{p}{2}} \vee \frac{2^{\frac{p}{2}}}{(mn)^{\frac{p}{2}}} \right],
\end{align*}
for some $C_4 > 0$.

Then by~\Cref{lem:errorreduction}, the total error is bounded by
\begin{align*}
    &2^{p-1} \left\{C_4 \cdot \left[\left(\frac{4\log (\frac{k}{n}+1)}{ml}\right)^{\frac{p}{2}}\vee \frac{2^{\frac{p}{2}}}{(mn)^{\frac{p}{2}}}\right]+ C_3 \cdot  \left[\left(\frac{t}{mnl}\right)^{\frac{p}{2}} \vee \frac{1}{(mn)^{\frac{p}{2}}}\right]\right\} \\
    = &O\left( \left(\frac{\log \left(\frac{k}{n}+1\right)}{ml}\right)^{\frac{p}{2}} \vee \frac{1}{(mn)^{\frac{p}{2}}}\right).
\end{align*}

\subsection{Proof of Proposition~\ref{lem:secsuccessivemain}: Analysis of The Protocol for Case~\ref{case4}}
\label{subsec:analysiscase4}

By the analysis in~\Cref{subsubsec:case3}, the estimation error for the reduced block distribution induced by the invocation of~$\mathrm{ASR}(\frac{m}{2},n,k_{a+1},l,p)$ is bounded by
\begin{align*}
    C_5 \cdot \left[ \left(\frac{\log \left(\frac{k_{a+1}}{n}+1\right)}{ml}\right)^{\frac{p}{2}} \vee \frac{1}{(mn)^{\frac{p}{2}}} \right] \leq \frac{C_5}{m^{\frac{p}{2}}},
\end{align*}
for some $C_5>0$.

We have $k_{u+1} \geq k_{a+1} > n$ and $\frac{l_0}{l} = 1 > \frac{1}{n}$. Then by~\Cref{lem:errorsub}, the estimation error for the conditional distribution induced by the $u$-th invocation of the subroutine~$\mathrm{ASRSub}(m_u, n, k_u, l,l_0,p)$ is bounded by
\begin{equation}
      C_6 \cdot \left(\frac{k_{u+1}}{m_un}\right)^{\frac{p}{2}} \leq C_6 \left(\frac{\frac{k}{2^{u(l-1)}}}{\frac{m}{2^{u+2}}n}\right)^{\frac{p}{2}}
      = C_6 \left(\frac{2^{u+2}k}{(2^{l-1})^u mn}\right)^{\frac{p}{2}},
\end{equation}
for some $C_6 > 0$.

Then by~\Cref{lem:errorreduction} and $l \geq 4$, the total error is bounded by
\begin{align*}
      &2^{a(p-1)} \cdot    \frac{C_5}{m^{\frac{p}{2}}} + C_6 \sum_{u = 1}^a 2^{u(p-1)} \cdot \left(\frac{2^{u+2}k}{(2^{l-1})^u mn}\right)^{\frac{p}{2}}
      \\
      \leq & 2  \left(\frac{k}{n(2^l-1)}\right)^{\frac{2(p-1)}{l}} \cdot \frac{C_5}{m^{\frac{p}{2}}}+  2^{3p}C_6 \left(\frac{k}{2^lmn}\right)^{\frac{p}{2}}
      \\
      = &O\left( \left(\frac{k}{2^lmn}\right)^{\frac{p}{2}} \right).
\end{align*}

%% file: A3.tex
\section{The Non-interactive Protocol for the TV Loss}
\label{sec:tvprotocol}
Consider the estimation problem under the TV loss, i.e. $p = 1$. In this section, we show that a uniform resource allocation plan is sufficient in this case, thanks to the error bound~\eqref{eq:errorreductionTV}. 
The advantage of the uniform allocation plan  is obvious, since there is no need for the decoder to send any message to the encoders. Hence a non-interactive protocol is immediate induced, only by changing~\eqref{eq:defrs} to
\begin{equation}
\label{eq:defrs'}
    r(s) = \frac{1}{t}
\end{equation}
in the successive refinement subroutine $\mathrm{ASRSub}(m',n,k,l,l_0, 1)$ in~\Cref{subsec:subroutine}.

To show~\Cref{thm:upperboundTV}, it remains to show the error bound in the following proposition.

\begin{proposition}
\label{lem:sectvprotocolmain}
For any $\bm{p}_W \in \Delta_{\mathcal{W}}$, the non-interactive protocol $\mathrm{ASR}(m,n,k,l,1)$ outputs an estimate $\hat{\bm{p}}_W$ satisfying,
\begin{enumerate}[1.]
\item \label{case1'} 
if $k \leq n$,  $m(l \wedge k) > 1000 k \log m \log n$, then $\mathbb{E}[\Vert \hat{\bm{p}}_W-\bm{p}_W \Vert_p^p] =  O\left(\sqrt{\frac{k^2}{mnl}} \vee \sqrt{\frac{k}{mn}} \right)$;

\item \label{case3'} if $n < k \leq (2^l-1) \cdot n$, $l \geq 2$ and $m(l \wedge n) >2000  n \log m \log n$, 
then $\mathbb{E}[\Vert \hat{\bm{p}}_W-\bm{p}_W \Vert_p^p] =  O\left( \sqrt{\frac{k\log \left(\frac{k}{n}+1\right)}{ml}} \vee \sqrt{\frac{k}{mn}} \right)$;

\item \label{case4'} if $k >(2^l-1) \cdot n$, $l \geq 4$ and $m(l \wedge n) > 4000 n \log m \log n$, then $\mathbb{E}[\Vert \hat{\bm{p}}_W-\bm{p}_W \Vert_p^p] =  O\left( \sqrt{\frac{k^2}{2^lmn}}\right)$.

\end{enumerate}
\end{proposition}

\subsection{Error Analysis of the Subroutine for $p = 1$}

First, the estimation error induced by the subroutine $\mathrm{ASRSub}(m',n,k,l,l_0,1)$ is described in the following lemma.

\begin{lemma}
\label{lem:errorsub'}
We have 
\begin{equation}
    \sum_{s = 1}^t  \mathbb{E}[p_B(s)\Vert\hat{\bm{p}}_s - \bm{p}_s\Vert_{\mathrm{TV}}]  = O\left( \sqrt{ \frac{ t}{m'}\left(1\vee\frac{t}{n} \right) \cdot \left(\frac{l_0}{l} \vee \frac{1}{n}\right)}  \right).
\end{equation}
\end{lemma}

\begin{IEEEproof}
If $m'n_0r(s) = \frac{m'n_0}{t} \leq 4$, since $\Vert\hat{\bm{p}}_s - \bm{p}_s\Vert_{\mathrm{TV}} \leq 2$, then
$$
p_B(s)\Vert\hat{\bm{p}}_s - \bm{p}_s\Vert_{\mathrm{TV}} \leq 2p_B(s) \leq 4\sqrt{ \frac{p_B(s)^2 t}{m'n_0}} \leq 4\sqrt{ \frac{p_B(s)^2 k}{m'n_0}}.
$$

Otherwise, we have $m'n_0r(s) = \frac{m'n_0}{t} > 4$, hence $N_s = \Theta \left( m'n_0r(s) \right) = \Theta \left( \frac{m'n_0}{t}\right)$.
Then $\tilde{W}^{s}_{u}$ for $u = 1,...,N_s$ are i.i.d. random variables with
\begin{equation}
    q_s \triangleq \mathbb{P}[\tilde{W}^{s}_{u} \neq \emptyset|\hat{\bm{p}}_B]  = \Theta \left(p_B(s) \left\lfloor \frac{n}{\lceil n_0 r(s) \rceil} \right\rfloor \wedge 1\right) =  \Theta \left(p_B(s) \left\lfloor \frac{n}{\lceil n_0/t \rceil} \right\rfloor \wedge 1\right). 
\end{equation}
Then we can establish the following lemma.

\begin{lemma}
$\mathbb{E}[p_B(s)\Vert\hat{\bm{p}}_s - \bm{p}_s\Vert_{\mathrm{TV}}]  \leq C \mathbb{E}\left[ \sqrt{\frac{ p_B(s)k}{m'nt} \vee \frac{ p_B(s)k}{m'nn_0 } \vee \frac{p_B(s)^2 k}{m'n_0}}\right]$ for some $C>0$.
\end{lemma}

\begin{IEEEproof}
By the Chernoff bound, we have 
\begin{equation}
    \mathbb{P}\left[N_s' \geq \frac{N_s q_s}{2}\Big| \hat{\bm{p}}_B\right] \leq \exp \left(-\frac{N_s q_s}{8} \right).
\end{equation}
And conditional on the event $\{\tilde{W}^{s}_{u} \neq \emptyset \}$, the distribution of $\tilde{W}^{s}_{u}$ is $\bm{p}_s$. 
By the Cauchy-Schwarz inequality and $p_s(w) \in [0,1]$,
\begin{align*}
\mathbb{E}\left[\Vert\hat{\bm{p}}_s - \bm{p}_s\Vert_{\mathrm{TV}} \Big| N_s' \geq \frac{N_s q_s}{2}\right] \leq \sqrt{|\mathcal{W}_s| \cdot \mathbb{E}\left[\Vert\hat{\bm{p}}_s - \bm{p}_s\Vert_{2}^2 \Big| N_s' \geq \frac{N_s q_s}{2}\right]} = O\left( \sqrt{\frac{|\mathcal{W}_s|}{N_s q_s}} \right). 
\end{align*}
Since $\Vert\hat{\bm{p}}_s - \bm{p}_s\Vert^2 \leq 2$, we have
\begin{align*}
\mathbb{E}[\Vert\hat{\bm{p}}_s - \bm{p}_s\Vert^2|\hat{\bm{p}}_B] \leq 2\exp \left(-\frac{N_s q_s}{8} \right)+ O\left( \sqrt{\frac{|\mathcal{W}_s|}{N_s q_s}} \right) = O\left( \sqrt{\frac{|\mathcal{W}_s|}{N_s q_s}} \right). 
\end{align*}

Since $n_0 \leq n$, we have $\lceil \frac{n_0}{t}\rceil \leq n$ and $ \frac{n}{\lceil n_0/t \rceil} \geq 1$. Hence there exists some $C>0$, such that
\begin{align*}
    & \mathbb{E}[p_B(s)\Vert\hat{\bm{p}}_s - \bm{p}_s\Vert_{\mathrm{TV}}] 
    \leq C \mathbb{E}\left[\sqrt{ \frac{p_B(s)^2 \frac{k}{t}}{\frac{m'n_0}{t} q_s} }\right]
    \\
    = & C\mathbb{E}\left[ \sqrt{\frac{p_B(s)^2k}{m'n_0 \left(p_B(s) \lfloor \frac{n}{\lceil n_0/t \rceil} \rfloor \wedge 1\right)} }\right]
    \\
    = &C \mathbb{E}\left[ \sqrt{ \frac{p_B(s)k}{m'n_0 \left( \frac{n}{\lceil n_0/t \rceil} \right)} \vee \frac{p_B(s)^2 k}{m'n_0} }\right]
    \\
    = &C \mathbb{E}\left[ \sqrt{\frac{ p_B(s)k}{m'nt} \vee \frac{ p_B(s)k}{m'nn_0 } \vee \frac{p_B(s)^2 k}{m'n_0}}\right],
\end{align*}
completing the proof.
\end{IEEEproof}

In both cases, we can take the expectation and obtain that 
\begin{align*}
    \mathbb{E}[p_B(s)\Vert\hat{\bm{p}}_s - \bm{p}_s\Vert_{\mathrm{TV}}] \leq C' \mathbb{E}\left[ \sqrt{\frac{ p_B(s)k}{m'nt} \vee \frac{ p_B(s)k}{m'nn_0 } \vee \frac{p_B(s)^2 k}{m'n_0}}\right],
\end{align*}
for some $C'>0$.

Finally, take the sum over $s$ and use the Cauchy-Schwarz inequality, then 
\begin{align*}
    &\sum_{s = 1}^t  \mathbb{E}[p_B(s)\Vert\hat{\bm{p}}_s - \bm{p}_s\Vert_{\mathrm{TV}}]
    \\
    \leq &C' \sum_{s = 1}^t \mathbb{E}\left[\sqrt{\frac{ p_B(s)k}{m'nt} \vee \frac{ p_B(s)k}{m'nn_0 } \vee \frac{p_B(s)^2 k}{m'n_0}}\right]
    \\
    = &O\left(\sqrt{ \frac{k}{m'n_0} \vee \frac{kt}{m'nn_0} }\right)
    \\
    = &O\left( \sqrt{ \frac{k}{m'}\left(1\vee\frac{t}{n} \right) \cdot \left(\frac{l_0}{l} \vee \frac{1}{n}\right)}  \right),
\end{align*}
which completes the proof of~\Cref{lem:sectvprotocolmain}.

\end{IEEEproof}

\subsection{Error Analysis of the Non-Interactive Protocol}
We complete the proof of~\Cref{lem:sectvprotocolmain} in this subsection.

\subsubsection{Error Analysis for the Base Case~\ref{case1'}}
\label{subsubsec:base'}

Since the protocol for $p = 1$ is the same as that for $p = 2$, then by the Cauchy-Schwarz inequality and the analysis in~\Cref{sec:base} we have
\begin{align*}
    &\mathbb{E}[\Vert\hat{\bm{p}}_W-\bm{p}_W\Vert_{\mathrm{TV}}] \leq \sqrt{k \mathbb{E}[\Vert\hat{\bm{p}}_W-\bm{p}_W\Vert_2^2]} \preceq \sqrt{\frac{k^2}{mnl}}  \vee \sqrt{\frac{k}{mn}}.
\end{align*}

\subsubsection{Error Analysis for Case~\ref{case3'}}
\label{subsubsec:case3'}

By the analysis in~\Cref{subsubsec:base'}, the estimation error for the reduced block distribution is bounded by
\begin{align*}
     C_3\cdot \left(\sqrt{\frac{t^2}{mnl}}  \vee \sqrt{\frac{t}{mn}}\right),
\end{align*}
for some $C_3 > 0$.

By~\Cref{lem:errorsub'}, the estimation error for the conditional distribution induced by the invoking of the subroutine $\mathrm{ASRSub}(\frac{m}{2},n,k,l,l_0,1)$ is bounded by
\begin{align*}
    C_4  \sqrt{\frac{k\left(\frac{l_0}{l} \vee \frac{1}{n}\right)}{\frac{m}{2}}} = C_4 \sqrt{\frac{2k\left(\frac{l_0}{l} \vee \frac{1}{n}\right)}{m}} = C_4 \left( \sqrt{\frac{2l_0k}{ml}}  \vee \sqrt{\frac{2k}{mn}}\right) \leq  C_4 \cdot\left(\sqrt{\frac{4k\log (\frac{k}{n}+1)}{ml}} \vee \sqrt{\frac{2k}{mn}}\right),
\end{align*}
for some $C_4 > 0$.

Then by~\eqref{eq:errorreductionTV}, the total error is bounded by
\begin{align*}
    & C_3 \cdot \left(\sqrt{\frac{t^2}{mnl}} \vee \sqrt{\frac{t}{mn}}\right) + C_4 \cdot \left(\sqrt{\frac{4k\log (\frac{k}{n}+1)}{ml}} \vee \sqrt{\frac{k}{mn}}\right) = O\left( \sqrt{\frac{k\log \left(\frac{k}{n}+1\right)}{ml}} \vee \sqrt{\frac{k}{mn}}\right).
\end{align*}

\subsubsection{Error Analysis for Case~\ref{case4'}}
\label{subsubsec:case4'}

By the analysis in~\Cref{subsubsec:case3'}, the estimation error for the reduced block distribution induced by the invocation of~$\mathrm{ASR}(\frac{m}{2},n,k_{a+1},l,1)$ is bounded by
\begin{align*}
    C_5 \cdot \left(\sqrt{\frac{k_{a+1}\log \left(\frac{k_{a+1}}{n}+1\right)}{ml}} \vee \sqrt{\frac{k_{a+1}}{mn}}\right) \leq C_5 \cdot  \sqrt{\frac{k_{a+1}}{m}},
\end{align*}
for some $C_5>0$.

We have $k_{u+1} \geq k_{a+1} > n$ and $\frac{l_0}{l} = 1 > \frac{1}{n}$. Then by~\Cref{lem:errorsub'}, the estimation error for the conditional distribution induced by the $u$-th invocation of the subroutine~$\mathrm{ASRSub}(m_u, n, k_u, l,l_0,1)$ is bounded by
\begin{equation}
      C_6 \cdot \sqrt{\frac{k_{u+1}\cdot k_u}{m_un}} \leq C_6 \sqrt{\frac{\frac{k}{2^{u(l-1)}} \cdot k}{\frac{m}{2^{u+2}}n}}
      = C_6 \sqrt{\frac{2^{u+2}k^2}{(2^{l-1})^u mn}},
\end{equation}
for some $C_6 > 0$.

Then by~\eqref{eq:errorreductionTV} and $l \geq 4$, the total error is bounded by
\begin{align*}
      C_5  \cdot \sqrt{\frac{k_{a+1}}{m}} + C_6 \sum_{u = 1}^a  \sqrt{\frac{2^{u+2}k^2}{(2^{l-1})^u mn}} \leq C_5 \cdot \sqrt{\frac{k}{m}}+  8C_6 \sqrt{\frac{k^2}{2^lmn}} = O\left( \sqrt{\frac{k^2}{2^lmn}} \right).
\end{align*}

%% file: A4.tex
\section{Proof of Proposition~\ref{lem:secmultiplesamplemain}: Error Analysis for the Protocol in Section~\ref{sec:multiplesample}}
\label{sec:pfmultiplesample}

To complete the proof of~\Cref{lem:secmultiplesamplemain}, it suffices to show that $\mathbb{E}[\Vert\hat{\bm{p}}^3_W-\bm{p}_W\Vert_p^p]  = O\left(\frac{ 1}{(mn_0)^{\frac{p}{2}}n^{\frac{p}{2}-1}}\right)$.

We can obtain the following preliminary results, characterizing the estimation errors for the first and the second step. 
The proof is derived from~\eqref{eq:step1error}, similar to the proof of~\Cref{lem:crudeerrorprobability'} but simpler.
\begin{equation}
\label{eq:testingerror}
    \mathbb{P}\left[\hat{p}^1_W(w) \leq \frac{p_W(w)}{2} \right] = O\left(\frac{1}{(mn_0p_W(w))^{\frac{p}{2}}}\right).
\end{equation}
By~\eqref{eq:baseerror} in the proof of~\Cref{lem:secbasemain}, we have 
\begin{equation}
\label{eq:step2error}
    \mathbb{E}\left[|p_W(w)-\hat{p}^3_{W'}(w)|^p| w \in \mathcal{W}'\right]  = O\left(\frac{1}{(mnl)^{\frac{p}{2}}} + \frac{p_W(w)^{\frac{p}{2}}}{(mn)^{\frac{p}{2}}}\right).
\end{equation}

Note that
\begin{align*}
    \mathbb{E}[\Vert\hat{\bm{p}}^3_W-\bm{p}_W\Vert_p^p] \leq \sum_{w: p_W(w) \leq \frac{4}{n}}\mathbb{E}\left[|p_W(w)-\hat{p}^3_{W}(w)|^p\right] + \sum_{w: p_W(w) > \frac{4}{n}}\mathbb{E}\left[|p_W(w)-\hat{p}^3_{W}(w)|^p\right].
\end{align*}
It suffices to bound the above two terms separately.

If $p_W(w) \leq \frac{4}{n}$, then by the error bounds~\eqref{eq:step1error} (applied to $\hat{\bm{p}}^2_W$) and ~\eqref{eq:step2error}, we have
\begin{align*}
    \mathbb{E}\left[|p_W(w)-\hat{p}^3_{W}(w)|^p\right] = &\mathbb{E}\left[\mathds{1}_{w \in \mathcal{W}'}|p_W(w)-\hat{p}^3_{W'}(w)|^p\right] + \mathbb{E}\left[\mathds{1}_{w \notin \mathcal{W}'}|p_W(w)-\hat{p}^2_{W}(w)|^p\right]
    \\
    \leq &\mathbb{P}[w \in \mathcal{W}']\mathbb{E}\left[|p_W(w)-\hat{p}^3_{W'}(w)|^p | w \in \mathcal{W}'\right] + \mathbb{E}\left[|p_W(w)-\hat{p}^2_{W}(w)|^p\right]
    \\
    \leq &O\left(\frac{\mathbb{P}[w \in \mathcal{W}']}{(mnl)^{\frac{p}{2}}} + \frac{p_W(w)^{\frac{p}{2}}}{(mn)^{\frac{p}{2}}}\right) + O \left( \left(\frac{p_W(w)}{mn_0}\right)^{\frac{p}{2}}\right)
    \\
    = &O\left(\frac{\mathbb{P}[w \in \mathcal{W}']}{(mnl)^{\frac{p}{2}}} + \frac{p_W(w)}{(mn_0)^{\frac{p}{2}}n^{\frac{p}{2}-1}}\right).
\end{align*}
Take the summation and note that $|\mathcal{W}'| \leq n$, then
\begin{equation}
\label{eq:sumless}
\begin{aligned}
    \sum_{w: p_W(w) \leq \frac{4}{n}}\mathbb{E}\left[|p_W(w)-\hat{p}^3_{W}(w)|^p\right] \leq &O \left(\sum_{w: p_W(w) \leq \frac{4}{n}} \frac{\mathbb{P}[w \in \mathcal{W}']}{(mnl)^{\frac{p}{2}}} + \frac{p_W(w)}{(mn_0)^{\frac{p}{2}}n^{\frac{p}{2}-1}} \right)
    \\
    \leq &O \left(\frac{\mathbb{E}[|\mathcal{W}'|]}{(mnl)^{\frac{p}{2}}} + \frac{1}{(mn_0)^{\frac{p}{2}}n^{\frac{p}{2}-1}} \right)
    = O \left( \frac{1}{(mn_0)^{\frac{p}{2}}n^{\frac{p}{2}-1}} \right).
\end{aligned}
\end{equation}

If $p_W(w) > \frac{4}{n}$, then $\mathbb{P}[w \notin \mathcal{W}'] \leq \mathbb{P}\left[\hat{p}^1_W(w) \leq \frac{p_W(w)}{2} \right]$. By~\eqref{eq:step1error} (applied to $\hat{\bm{p}}^2_W$), \eqref{eq:testingerror} and~\eqref{eq:step2error}, we have
\begin{align*}
    \mathbb{E}\left[|p_W(w)-\hat{p}^3_{W}(w)|^p\right] = &\mathbb{E}\left[\mathds{1}_{w \in \mathcal{W}'}|p_W(w)-\hat{p}^3_{W'}(w)|^p\right] + \mathbb{E}\left[\mathds{1}_{w \notin \mathcal{W}'}|p_W(w)-\hat{p}^2_{W}(w)|^p\right]
    \\
    \leq &\mathbb{E}\left[|p_W(w)-\hat{p}^3_{W'}(w)|^p | w \in \mathcal{W}'\right] + \mathbb{P}[w \notin \mathcal{W}'] \cdot \mathbb{E}\left[|p_W(w)-\hat{p}^2_{W}(w)|^p\right]
    \\
    \leq &O\left(\frac{1}{(mnl)^{\frac{p}{2}}} + \frac{p_W(w)^{\frac{p}{2}}}{(mn)^{\frac{p}{2}}}\right) + O\left(\frac{1}{(mn_0 p_W(w))^{\frac{p}{2}}} \cdot \left(\frac{p_W(w)}{mn_0}\right)^{\frac{p}{2}}\right)
    \\
    = &O\left(\frac{1}{(mnn_0)^{\frac{p}{2}}} + \frac{p_W(w)^{\frac{p}{2}}}{(mn)^{\frac{p}{2}}}\right),
\end{align*}
where the last step is since $mn_0 \geq \frac{ml}{4\log k} > 1000 n$.
Take the summation and note that $|\{w: p_W(w) > \frac{4}{n}\}| \leq n$, we have
\begin{equation}
\label{eq:sumgreater}
\begin{aligned}
    &\sum_{w: p_W(w) > \frac{4}{n}}\mathbb{E}\left[|p_W(w)-\hat{p}^3_{W}(w)|^p\right] \leq O \left(\sum_{w: p_W(w) > \frac{4}{n}} \frac{1}{(mnn_0)^{\frac{p}{2}}} + \frac{p_W(w)^{\frac{p}{2}}}{(mn)^{\frac{p}{2}}} \right)
    = O \left( \frac{1}{(mn_0)^{\frac{p}{2}}n^{\frac{p}{2}-1}} \right),
\end{aligned}
\end{equation}
where the last step is since $n_0 
 = \lfloor \frac{l}{\lceil \log k \rceil} \rfloor \leq n^{\frac{2}{p}}$.
Combining~\eqref{eq:sumless} and~\eqref{eq:sumgreater}, we complete the proof of~\Cref{lem:secmultiplesamplemain}.

%% file: A5.tex
\section{Proof of Proposition~\ref{lem:sectightbudgetmain}: Error Analysis for the protocol in Section~\ref{sec:tightbudget}}
\label{sec:pftightbudget}

\subsection{Error Analysis for the Protocol in Section~\ref{subsec:thresholding}}
It suffices to show that $\mathbb{E}[\Vert\hat{\bm{p}}^3_W-\bm{p}_W\Vert_p^p]  = O\left(\frac{1}{(mn_0)^{\frac{p}{2}} (ml)^{\frac{p}{2}-1}}\right)$.

We first give the following preliminary results, characterizing the estimation error for the first step. 
The proof is derived from~\eqref{eq:step1error}, similar to the proof of~\Cref{lem:crudeerrorprobability'} but simpler.
\begin{equation}
\label{eq:testingerror1}
    \mathbb{P}\left[\hat{p}^1_W(w) \leq \frac{p_W(w)}{2} \right] \leq O\left(\frac{1}{(mn_0p_W(w))^{\frac{p}{2}}}\right).
\end{equation}

Note that
\begin{align*}
    \mathbb{E}[\Vert\hat{\bm{p}}^3_W-\bm{p}_W\Vert_p^p] \leq \sum_{w: p_W(w) \leq \frac{4}{ml}}\mathbb{E}\left[|p_W(w)-\hat{p}^3_{W}(w)|^p\right] + \sum_{w: p_W(w) > \frac{4}{ml}}\mathbb{E}\left[|p_W(w)-\hat{p}^3_{W}(w)|^p\right].
\end{align*}
It suffices to bound the two terms separately. If $p_W(w) \leq \frac{4}{ml}$, then by~\eqref{eq:step1error} (applied to~$\hat{\bm{p}}^2_{W'}$),
\begin{align*}
    \mathbb{E}\left[|p_W(w)-\hat{p}^3_{W}(w)|^p\right] = &\mathbb{E}\left[\mathds{1}_{w \in \mathcal{W}'}|p_W(w)-\hat{p}^2_{W}(w)|^p\right] + \mathbb{E}\left[\mathds{1}_{w \notin \mathcal{W}'}p_W(w)^p\right]
    \\
    \leq &\mathbb{P}[w \in \mathcal{W}'] \cdot \mathbb{E}\left[|p_W(w)-\hat{p}^2_{W}(w)|^p\right] + p_W(w)^p
    \\
    = &O\left(\mathbb{P}[w \in \mathcal{W}']\cdot \left(\frac{p_W(w)}{mn_0}\right)^{\frac{p}{2}}\right) + p_W(w)^p
    \\
    = &O\left(\mathbb{P}[w \in \mathcal{W}']\cdot \left(\frac{1}{m^2n_0l}\right)^{\frac{p}{2}} + \frac{p_W(w)}{(ml)^{p-1}}\right).
\end{align*}
Take the summation and note that $|\mathcal{W}'| \leq ml$, then
\begin{equation}
\label{eq:sumless1}
\begin{aligned}
    \sum_{w: p_W(w) \leq \frac{4}{ml}}\mathbb{E}\left[|p_W(w)-\hat{p}^3_{W}(w)|^p\right] \leq &O \left(\sum_{w: p_W(w) \leq \frac{4}{ml}} \mathbb{P}[w \in \mathcal{W}']\cdot \left(\frac{1}{m^2n_0l}\right)^{\frac{p}{2}} + \frac{p_W(w)}{(ml)^{p-1}} \right)
    \\
    \leq &O \left(\frac{\mathbb{E}[|\mathcal{W}'|]}{(m^2n_0l)^{\frac{p}{2}}} + \frac{1}{(ml)^{p-1}} \right)
    = O \left( \frac{1}{(mn_0)^{\frac{p}{2}}(ml)^{\frac{p}{2}-1}} \right).
\end{aligned}
\end{equation}

If $p_W(w) > \frac{4}{ml}$, then $\mathbb{P}[w \notin \mathcal{W}'] \leq \mathbb{P}\left[\hat{p}^1_W(w) \leq \frac{p_W(w)}{2} \right]$. By~\eqref{eq:step1error} (applied to~$\hat{\bm{p}}^2_W$) and ~\eqref{eq:testingerror1}, we have
\begin{align*}
    \mathbb{E}\left[|p_W(w)-\hat{p}^3_{W}(w)|^p\right] = &\mathbb{E}\left[\mathds{1}_{w \in \mathcal{W}'}|p_W(w)-\hat{p}^2_{W}(w)|^p\right] + \mathbb{E}\left[\mathds{1}_{w \notin \mathcal{W}'} p_W(w)^p\right]
    \\
    \leq &\mathbb{E}\left[|p_W(w)-\hat{p}^2_{W}(w)|^p\right] + \mathbb{P}[w \notin \mathcal{W}'] \cdot p_W(w)^p
    \\
    \leq &O\left( \left(\frac{p_W(w)}{mn_0}\right)^{\frac{p}{2}} \right) + p_W(w)^p \cdot O\left(\frac{1}{(mn_0p_W(w))^{\frac{p}{2}}}\right)
    \\
    = &O\left( \left(\frac{p_W(w)}{mn_0}\right)^{\frac{p}{2}} \right).
\end{align*}
Taking the summation and noting that $|\{w: p_W(w) > \frac{4}{ml}\}| \leq ml$, by the H\"{o}lder's inequality we have
\begin{equation}
\label{eq:sumgreater1}
\begin{aligned}
    &\sum_{w: p_W(w) > \frac{4}{ml}}\mathbb{E}\left[|p_W(w)-\hat{p}^3_{W}(w)|^p\right] \leq O \left(\sum_{w: p_W(w) > \frac{4}{ml}} \left(\frac{p_W(w)}{mn_0}\right)^{\frac{p}{2}}  \right)
    = O \left( \frac{1}{(mn_0)^{\frac{p}{2}}(ml)^{\frac{p}{2}-1}} \right).
\end{aligned}
\end{equation}
Combining~\eqref{eq:sumless1} and~\eqref{eq:sumgreater1}, we complete the proof.

\subsection{Error Analysis for the Protocol in Section~\ref{subsec:combining}}

It remains to show that $\mathbb{E}[\Vert\hat{\bm{p}}^3_W-\bm{p}_W\Vert_p^p]  = O\left(\frac{ml}{(k'\wedge mn_0  )^{p}}\right)$.

We first give the following preliminary results, characterizing the estimation error for the first step. 
The proof is derived from~\eqref{eq:step1error}, similar to the proof of~\Cref{lem:crudeerrorprobability'} (where $p$ in~\Cref{lem:crudeerrorprobability'} is replaced by $2p$) but simpler.
\begin{equation}
\label{eq:testingerror2}
    \mathbb{P}\left[\hat{p}^1_W(w) \leq \frac{p_W(w)}{2} \right] \leq O\left(\frac{1}{(mn_0p_W(w))^{p}}\right).
\end{equation}
By~\eqref{eq:baseerror} in the proof of~\Cref{lem:secbasemain}, we have 
\begin{equation}
\label{eq:step2error2}
    \mathbb{E}\left[|p_W(w)-\hat{p}^2_{W'}(w)|^p| w \in \mathcal{W}'\right]  = O\left(\frac{1}{(mnl)^{\frac{p}{2}}} + \frac{p_W(w)^{\frac{p}{2}}}{(mn)^{\frac{p}{2}}}\right).
\end{equation}

Note that
\begin{align*}
    \mathbb{E}[\Vert\hat{\bm{p}}^3_W-\bm{p}_W\Vert_p^p] \leq \sum_{w: p_W(w) \leq \frac{4}{k'}}\mathbb{E}\left[|p_W(w)-\hat{p}^3_{W}(w)|^p\right] + \sum_{w: p_W(w) > \frac{4}{k'}}\mathbb{E}\left[|p_W(w)-\hat{p}^3_{W}(w)|^p\right].
\end{align*}
It suffices to bound the two terms separately. If $p_W(w) \leq \frac{4}{k'}$, then by~\eqref{eq:step2error2} (applied to~$\hat{\bm{p}}^2_W$), we have
\begin{align*}
    \mathbb{E}\left[|p_W(w)-\hat{p}^3_{W}(w)|^p\right] = &\mathbb{E}\left[\mathds{1}_{w \in \mathcal{W}'}|p_W(w)-\hat{p}^2_{W'}(w)|^p\right] + \mathbb{E}\left[\mathds{1}_{w \notin \mathcal{W}'}p_W(w)^p\right]
    \\
    \leq &\mathbb{P}[w \in \mathcal{W}']\mathbb{E}\left[|p_W(w)-\hat{p}^2_{W'}(w)|^p | w \in \mathcal{W}'\right] + p_W(w)^p
    \\
    \leq &O\left(\frac{\mathbb{P}[w \in \mathcal{W}']}{(mnl)^{\frac{p}{2}}} + \frac{p_W(w)^{\frac{p}{2}}}{(mn)^{\frac{p}{2}}}\right) + O \left( \frac{p_W(w)}{k'^{p-1}}\right) 
    \\
    = &O\left(\frac{\mathbb{P}[w \in \mathcal{W}']}{(mnl)^{\frac{p}{2}}} + \frac{p_W(w)}{k'^{p-1}}\right).
\end{align*}
Take the summation and note that $|\mathcal{W}'| \leq k'$, then 
\begin{equation}
\label{eq:sumless2}
\begin{aligned}
    \sum_{w: p_W(w) \leq \frac{4}{k'}}\mathbb{E}\left[|p_W(w)-\hat{p}^3_{W}(w)|^p\right] \leq &O \left(\sum_{w: p_W(w) \leq \frac{4}{k'}} \frac{\mathbb{P}[w \in \mathcal{W}']}{(mnl)^{\frac{p}{2}}} + \frac{p_W(w)}{k'^{p-1}} \right)
    \\
    \leq &O \left(\frac{\mathbb{E}[|\mathcal{W}'|]}{(mnl)^{\frac{p}{2}}} + \frac{1}{k'^{p-1}} \right)
    = O \left( \frac{1}{k'^{p-1}} \right).
\end{aligned}
\end{equation}

If $p_W(w) > \frac{4}{k'}$, then $\mathbb{P}[w \notin \mathcal{W}'] \leq \mathbb{P}\left[\hat{p}^1_W(w) \leq \frac{p_W(w)}{2} \right]$. By~\eqref{eq:testingerror2} and~\eqref{eq:step2error2}, we have
\begin{align*}
    \mathbb{E}\left[|p_W(w)-\hat{p}^3_{W}(w)|^p\right] = &\mathbb{E}\left[\mathds{1}_{w \in \mathcal{W}'}|p_W(w)-\hat{p}^2_{W'}(w)|^p\right] + \mathbb{E}\left[\mathds{1}_{w \notin \mathcal{W}'}p_W(w)^p\right]
    \\
    \leq &\mathbb{E}\left[|p_W(w)-\hat{p}^2_{W'}(w)|^p | w \in \mathcal{W}'\right] + \mathbb{P}[w \notin \mathcal{W}'] \cdot p_W(w)^p
    \\
    \leq &O\left(\frac{1}{(mnl)^{\frac{p}{2}}} + \frac{p_W(w)^{\frac{p}{2}}}{(mn)^{\frac{p}{2}}}\right) + O\left(\frac{1}{(mn_0 p_W(w))^{p}} \cdot p_W(w)^p\right)
    \\
    = &O\left(\frac{1}{(mn_0)^{p}} + \frac{p_W(w)^{\frac{p}{2}}}{(mn)^{\frac{p}{2}}}\right),
\end{align*}
where the last step is since $mn_0 = m\lfloor \frac{l}{\lceil \log k \rceil} \rfloor < ml < n$.
Take the summation and note that $|\{w: p_W(w) > \frac{4}{k'}\}| \leq k' < ml$, we have
\begin{equation}
\label{eq:sumgreater2}
\begin{aligned}
    &\sum_{w: p_W(w) > \frac{4}{k'}}\mathbb{E}\left[|p_W(w)-\hat{p}^3_{W}(w)|^p\right] \leq O \left(\sum_{w: p_W(w) > \frac{4}{k'}} \frac{1}{(mn_0)^{p}} + \frac{p_W(w)^{\frac{p}{2}}}{(mn)^{\frac{p}{2}}} \right)
    = O \left( \frac{ml}{(mn_0)^{p}} \vee \frac{1}{(mn)^{\frac{p}{2}}} \right).
\end{aligned}
\end{equation}
Combining~\eqref{eq:sumless2} and~\eqref{eq:sumgreater2}, we complete the proof.

%% file: A6.tex
\section{Proof of  Proposition~\ref{lem:seconesamplemain}: Error Analysis for the Protocol in Section~\ref{sec:onesample}}
\label{sec:pfonesample}

We can analyze the error of the estimate $\hat{\bm{p}}_W$ as follows. 
Note that for each $w \in \mathcal{W}$ and $i = 1,...,m$,
\begin{equation*}
    \mathbb{P}[h_i(w) = B_i] = p_W(w) + \frac{1}{2^l} \left(1-p_W(w)\right).
\end{equation*}
It is folklore that (cf. Theorem 4 in~\cite{Skorski2020}),
\begin{equation*}
\mathbb{E}\left[\left|\frac{\sum_{i = 1}^{m} \mathds{1}_{h_i(w) = B_i}}{m}-p_W(w)\right|^p\right] = O \left( \left(\frac{\mathbb{P}[h_1(w) = B_1]}{m}\right)^{\frac{p}{2}}\right) =  O \left( \left(\frac{p_W(w) \vee \frac{1}{2^l}}{m}\right)^{\frac{p}{2}}\right).
\end{equation*}
Then by~\eqref{eq:rescaling}, we have
\begin{equation*}
\mathbb{E}[|\hat{p}_W(w)-p_W(w)|^p] = O \left( \left(\frac{p_W(w) \vee \frac{1}{2^l}}{m}\right)^{\frac{p}{2}}\right)
\end{equation*}
as well. By taking the summation over all $w \in \mathcal{W}$, we complete the proof of~\Cref{lem:seconesamplemain}.

%% file: A7.tex
\section{Proof of Lemma~\ref{lem:lowerboundregular}}
\label{pflem:lowerboundregular}

For $p = 1$, we have the following lemma in~\cite{Acharya2021}.

\begin{lemma}[\cite{Acharya2021}, Theorem 1.1 \& 1.3]
\label{lem:lowerbound'}
\begin{enumerate}
    \item For $n \geq k \log k$ and $m > \left(\frac{k}{l}\right)^2$, $R(m,n,k,l,1) \succeq \sqrt{\frac{k^2}{mnl}} \wedge 1$.
    \item For $n \leq k \log k$ and $m > \left(\frac{k}{l}\right)^2$, $R(m,n,k,l,1) \succeq \sqrt{\frac{k}{ml\log k}} \wedge 1$.
    \item We always have $R(m,n,k,l,1) \succeq \sqrt{\frac{k^2}{mn2^l}} \wedge 1$.
\end{enumerate}
\end{lemma}

With the help of~\eqref{eq:1top}, the following three bounds is derived from three cases in~\Cref{lem:lowerbound'} respectively. 

\subsection{Proof of the First Bound} For $n \geq k \log k$ and $m > (\frac{k}{l})^2$ and $l \leq k$, we can obtain that $m > \frac{k}{l}$ and $mnl \geq k^2$. Then by 1) in~\Cref{lem:lowerbound'} and~\eqref{eq:1top},
\begin{equation*}
    R(m,n,k,l,p) \succeq \frac{k}{(mnl)^{\frac{p}{2}}}. 
\end{equation*}

\subsection{Proof of the Second Bound} If $m > (\frac{k}{l})^2$ and $l \leq k$, then $ml\log k \geq k$. Then by 2) in~\Cref{lem:lowerbound'} and~\eqref{eq:1top} we have
\begin{equation*}
    R(m,n,k,l,p) \succeq \frac{k^{1-\frac{p}{2}}}{(ml \log k)^{\frac{p}{2}}}.
\end{equation*}

Now let $p \geq 2$. Since $n \leq k \log k$ we have $k \geq \frac{n}{\log n}$.   We further have
\begin{equation*}
    R(m,n,k,l,p) \geq R(m,n,\lceil n/\log n \rceil,l,p) \succeq \frac{1}{(ml)^{\frac{p}{2}}n^{\frac{p}{2}-1}\log n}.
\end{equation*}
as long as $m > (\frac{\lceil n/\log n \rceil}{l})^2$ and $l \leq \lceil n/\log n \rceil$.

\subsection{Proof of the Third Bound} 

If $mn2^l \geq k^2$, then by 1) in~\Cref{lem:lowerbound'} and~\eqref{eq:1top} we have
\begin{equation*}
    R(m,n,k,l,p) \succeq \frac{k}{(mn2^l)^{\frac{p}{2}}}.
\end{equation*}